\documentclass[10pt,twocolumn,letterpaper]{article}

\usepackage[pagenumbers]{cvpr}

\usepackage{cuted} 

\usepackage{currfile} 

\usepackage{caption} 

\newif\ifrevisemark
\revisemarkfalse

\usepackage{algorithm}
\usepackage{pifont}
\usepackage{array}
\usepackage{multirow}
\newcolumntype{R}[1]{>{\raggedright\arraybackslash}p{#1}}

\usepackage{algpseudocode}
\usepackage{colortbl}
\definecolor{oursrow}{gray}{0.92}
\definecolor{labelrow}{gray}{0.95} 
\newcommand{\labelrowfirst}[1]{\multicolumn{1}{@{}>{\columncolor{labelrow}[0pt][\tabcolsep]}l}{#1}}
\newcommand{\labelrowlast}[1]{\multicolumn{1}{>{\columncolor{labelrow}[\tabcolsep][0pt]}c@{}}{#1}}

\usepackage{amsthm}
\newtheoremstyle{paperthm}{3pt}{3pt}{\itshape}{}{\bfseries}{.}{0.5em}{}
\theoremstyle{paperthm}

\newtheorem{lemma}{Lemma}
\AtEndPreamble{%
  \crefname{proposition}{Prop.}{Props.}%
  \Crefname{proposition}{Proposition}{Propositions}%
  \crefname{assumption}{Assumption}{Assumptions}%
  \Crefname{assumption}{Assumption}{Assumptions}%
  \crefname{lemma}{Lemma}{Lemmas}%
  \Crefname{lemma}{Lemma}{Lemmas}%
}

\providecommand{\argmax}{}\renewcommand{\argmax}{\operatorname*{arg\,max}}

\providecommand{\nrm}{}\renewcommand{\nrm}{\operatorname{nrm}}

\newcommand{\Praw}{P^{\mathrm{raw}}}
\newcommand{\Pbal}{P^{\mathrm{bal}}}
\newcommand{\Psel}{P^{\mathrm{sel}}}
\newcommand{\pzs}{p^{\mathrm{zs}}}
\newcommand{\pzsbar}{\bar p^{\mathrm{zs}}}
\newcommand{\Sw}{\Sigma_{\mathrm{w}}}
\newcommand{\Keff}{K_{\mathrm{eff}}}

\definecolor{cvprblue}{rgb}{0.21,0.49,0.74}
\usepackage[pagebackref,breaklinks,colorlinks,allcolors=cvprblue]{hyperref}
\usepackage[permil]{overpic}

\newtheorem{theorem}{Theorem}

\AtEndPreamble{%
  \crefname{figure}{Fig.}{Figs.}%
  \crefname{table}{Table}{Tables}%
  \crefname{equation}{Eq.}{Eqs.}%
}

\newcommand{\ours}[0]{\textsc{PACIL}\xspace}
\providecommand{\natexlab}[1]{}\renewcommand{\natexlab}[1]{}

\def\confName{CVPR}
\def\confYear{2027}

\title{Learning What the True Labels Would Teach: Posterior Analytic Class-Incremental Learning with Unlabeled Data}

\author{
 Wei Tang$^{1}$, \space Tong Wei$^{2}$, \space Jiahao Zhang$^{1}$, \space Min-Ling Zhang$^{2}$,  \space Lijie Hu$^{1}$,  \space Fakhri Karray$^{1}$
 \\
  $^1$Mohamed bin Zayed University of Artificial Intelligence (MBZUAI),   $^2$Southeast University \\
  {\tt\small \{wei.tang,jiahao.zhang,lijie.hu,fakhri.karray\}@mbzuai.ac.ae, \{weit,zhangml\}@seu.edu.cn}
}

\begin{document}
\maketitle
\begin{abstract}
In class-incremental learning with unlabeled data, each task brings only class names and unlabeled images, so CLIP's zero-shot predictions replace labels. Existing methods train on hard pseudo-labels, but a wrong one gives the true class zero probability, which self-training often confirms. We argue that each image should carry its label posterior, as an analytic classifier trained on posteriors equals the true-label one in expectation. The zero-shot prediction is not this posterior, as its class proportions confound name bias with true imbalance and it ignores other images. We propose Posterior Analytic Class-Incremental Learning ({\ours}), which corrects the class prior with a prior selection rule and refines each distribution by a fixed-prior von Mises-Fisher mixture on whitened features. These soft labels train the classifier and replace labels to select its fusion weight with zero-shot CLIP by closed-form leave-one-out cross-validation. We prove that posterior targets minimize the mean-square gap to the true-label prediction and that their argmax widens this gap only through ambiguous images. The fusion weight selected with soft labels loses at most their distance to the posteriors in expected correct decisions. {\ours} has the best average incremental accuracy among methods trained without labels on all 18 settings.
\end{abstract}

\section{Introduction}
\label{sec:intro}

\begin{figure}[!t]
\setlength{\abovecaptionskip}{0.cm}
    \centering
	\begin{overpic}[width=8.cm]{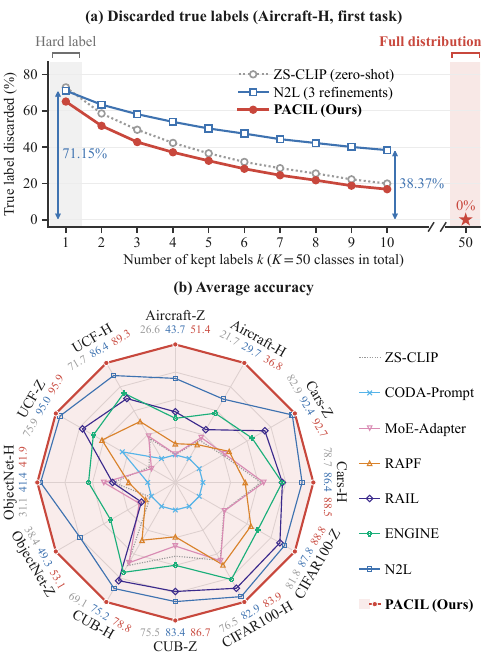}
    \end{overpic}
    \caption{Discarded true labels and average accuracy.
(a)~Training images (\%) whose true label is discarded when each keeps only its top-scored labels, on the first task of Aircraft (half-base), for zero-shot CLIP, N2L after three self-training refinements and {\ours}.
(b)~Average accuracy $\bar A$ (\%) of the methods of \cref{tab:main} on its zero-base (Z) and half-base (H) settings, each axis spanning the lowest (inner ring) to the highest value (outer ring). Numbers are the $\bar A$ of ZS-CLIP (gray), N2L (blue), and {\ours} (red).}
 \label{fig:pseudo_label_acc_and_teaser}
\end{figure}

Vision-language models such as CLIP~\cite{radford2021clip} classify images of any named class, so their zero-shot predictions can supervise models on unlabeled images. Pseudo-labeling methods reduce each prediction to one class or a candidate set~\cite{mirza2023lafter,zhang2024cpl}, while transductive methods keep the whole distribution~\cite{martin2024transductive,huang2026moon} but only to label the images they are fitted on, not future ones. What target each image should carry when a classifier is trained on such predictions remains open.

This question is critical in class-incremental learning (CIL)~\cite{zhou2024cilsurvey}, where a model learns disjoint classes in sequence and must recognize all seen classes without revisiting old data. In CIL with unlabeled data~\cite{li2026n2l,kuchibhotla2026afcil}, each task brings only its new class names and unlabeled images. This arises when categories change faster than annotation, as for retail products~\cite{griffioen2024retail} or camera-trap wildlife surveys, whose 3.2 million images would take over eight years of full-time work to label~\cite{norouzzadeh2018camera}. Earlier images may also be withheld for privacy~\cite{zhuang2022acil}, and each target is then fixed once its task ends. Existing methods, from CLIP-based CIL on zero-shot argmax labels~\cite{zheng2023zscl,xu2024rail,zhou2025engine,kuchibhotla2026afcil} to N2L~\cite{li2026n2l} with self-training, all use hard pseudo-labels. \Cref{fig:pseudo_label_acc_and_teaser}a measures how often the true label is discarded. A hard label discards it for most images, and keeping more top-scored labels lowers this rate slowly. N2L refines the hard labels three times, which may introduce confirmation bias~\cite{arazo2020pseudo}. With two to ten labels kept, it discards more true labels than zero-shot CLIP, as fitting a wrong label suppresses the true one. Only the full distribution discards no true label. The target of each image should therefore keep it and give the true label as much weight as the observations support. \looseness=-1

We argue that this target is the label posterior given all observations. Soft targets in distillation~\cite{hinton2015distilling} and expectation-maximization (EM)~\cite{nigam2000text} rest on the same idea, which becomes exact for analytic learning~\cite{zhuang2022acil,mcdonnell2023ranpac}, the exemplar-free learner we build on. Its solution is linear in the targets, so training on the posteriors yields the conditional mean of the classifier that the true labels would give. A hard label departs from this target on every ambiguous image, even when its argmax is correct. Because the posteriors also give the probability that each decision is correct, they can replace labels when choosing the weight that fuses the classifier with zero-shot CLIP. However, keeping the whole zero-shot distribution rather than its argmax is not enough. Its class proportions mix the name bias of CLIP with true imbalance~\cite{parashar2024neglected,zhu2023gla,tang2025unleashing} in shares that vary across datasets and tasks, and it ignores the other images of the task. Without labels or old images, three questions arise. (i) Which class proportions should the prior of each new task follow? (ii) How far should the images of a task revise each distribution when old classes are kept only as statistics? (iii) How much should predictions over all seen classes trust the analytic classifier rather than zero-shot CLIP?

We answer these questions with Posterior Analytic Class-Incremental Learning ({\ours}), whose soft labels estimate the posteriors. For (i), a prior selection rule tests on the images of each task whether the skew of its zero-shot class proportions is name bias or true imbalance, which text alone cannot separate. For (ii), the selected distribution is a fixed per-image prior of a von Mises-Fisher (vMF) mixture fitted by EM on features whitened with statistics of all seen classes, so the images revise each distribution but not the class proportions. For (iii), closed-form leave-one-out (LOO) cross-validation picks the fusion weight of each task with the most correct decisions expected under the soft labels. In our analysis, rounding the posteriors to their argmax adds exactly the squared change this causes in the prediction, a cost that comes only from ambiguous posteriors. The distance of the soft labels to the posteriors bounds both the extra prediction error and what the selected fusion weight can lose in expected correct decisions. Prior selection and EM are therefore designed to reduce this distance before a task ends and its targets are fixed.

Our contributions are as follows. We show that hard pseudo-labels discard plausible classes and that their errors persist because targets are fixed after each task, and we therefore cast labeling as posterior estimation. We prove that posteriors are mean-square optimal targets for analytic learning and that they yield a label-free fusion criterion whose loss is bounded by the soft-label error. We propose {\ours}, an exemplar-free method that estimates these posteriors and attains the best average incremental accuracy among label-free methods on every setting, especially where zero-shot CLIP is weakest (\cref{fig:pseudo_label_acc_and_teaser}b).

\begin{figure*}[!t]
    \centering
	\begin{overpic}[width=17.4cm]{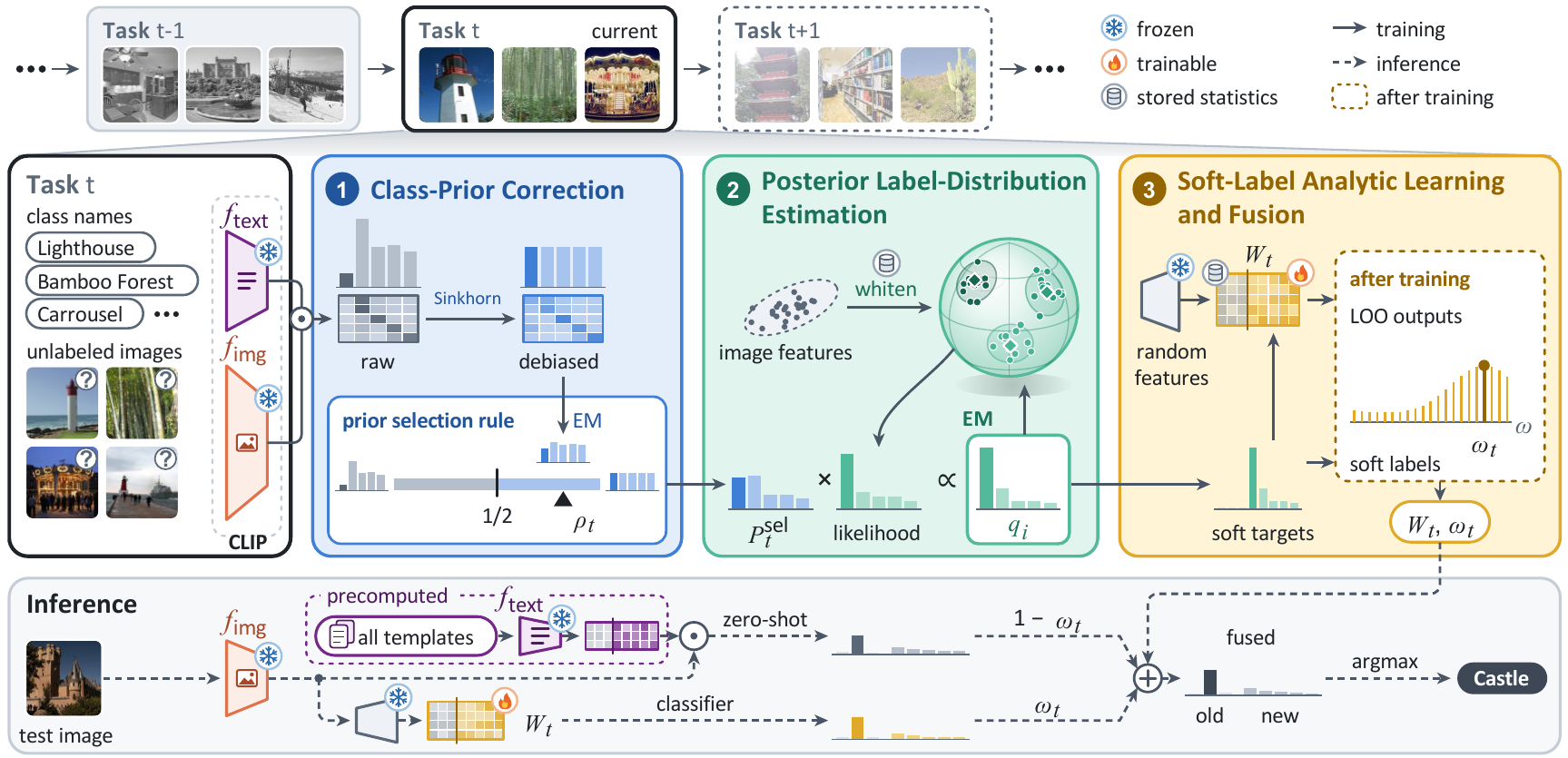}
    \end{overpic}
    \caption{Overview of {\ours} on task $t$.
(1)~Class-prior correction: the prior selection rule chooses $\Psel_t$ between the raw and the debiased zero-shot distribution.
(2)~Posterior label-distribution estimation: EM of a vMF mixture on whitened features keeps $\Psel_t$ as a fixed per-image prior and combines it with a tempered likelihood into the soft labels $q_i$.
(3)~Soft-label analytic learning and fusion: the analytic classifier $W_t$ is fitted to $q_i$, and closed-form LOO selects the weight $\omega_t$ that fuses it with zero-shot CLIP.}
 \label{fig:overview}
\end{figure*}

\section{Related Work}
\label{sec:related}

\paragraph{Class-Incremental Learning with CLIP.}
Most CLIP-based CIL methods tune CLIP or added modules by gradient descent on the labeled images of each task while limiting forgetting~\cite{zheng2023zscl,yu2024moe,huang2024rapf,zhou2025proof,zhou2025engine,luo2025lada}. Analytic learning replaces gradient training by a recursively updated ridge regression on frozen features, reaching the joint-training solution without exemplars~\cite{zhuang2022acil,zhuang2024dsal,zhuang2024gacl,mcdonnell2023ranpac}. RAIL~\cite{xu2024rail} and VILA~\cite{zhao2026vila} fuse such a learner with zero-shot CLIP by a fixed rule, and AnytimeCL~\cite{zhu2024anytimecl} weights a tuned model per class by labeled accuracy. All of these methods need labeled tasks. Class-incremental novel class discovery~\cite{roy2022incd,ma2024happy} and unsupervised CIL~\cite{sun2025dssd} learn from unlabeled images without class names, and name-only continual learning~\cite{prabhu2024nameonly} retrieves web images. N2L~\cite{li2026n2l} introduced CIL with unlabeled data and trains an analytic learner on zero-shot argmax labels refined by self-training. CrossWorld-CL~\cite{kuchibhotla2026afcil} tunes prompts and adapters on confident hard pseudo-labels in a similar setting, replaying retrieved ImageNet images. {\ours} trains an analytic learner on estimated label posteriors and selects its fusion weight with them.  \looseness=-1

\paragraph{Learning from Unlabeled Images with CLIP.}
Methods that adapt CLIP to unlabeled images differ in how much of its zero-shot distribution they keep. Pseudo-labeling methods keep one class per image~\cite{huang2022upl,li2023must,mirza2023lafter} or a candidate set~\cite{zhang2024cpl} and tune CLIP's prompts on them. Transductive methods keep the whole distribution and refine it over a set of test images, and test-time methods~\cite{dai2025freetta,zhou2025bca} do so over a stream.
Martin \etal~\cite{martin2024transductive} fit a Dirichlet mixture to CLIP's probability vectors and re-estimate the class proportions of each batch. TransCLIP~\cite{zanella2024transclip} fits a Gaussian mixture with shared covariance to the image features, regularized by a neighborhood graph and a KL term toward the zero-shot prediction. MOON~\cite{huang2026moon} uses vMF components whose mean directions shrink toward the text embeddings. Without graph terms, both multiply each zero-shot prediction by a class-conditional likelihood, the form of our E-step. Frolic~\cite{zhu2024frolic} removes an estimated label bias under a uniform test prior and fuses CLIP with a feature-space classifier at a weight equalizing their average confidences. These methods either predict the images they are fitted on or train on hard labels. Unlike Martin et al.~\cite{martin2024transductive}, TransCLIP~\cite{zanella2024transclip} and Frolic~\cite{zhu2024frolic}, {\ours} fixes the prior of each image from the prior selection rule and whitens features with the within class scatter of all seen classes.

\paragraph{Class Priors and Label Bias.}
Class priors are usually corrected with labels or a known source prior. Under label shift, Saerens \etal~\cite{saerens2002adjusting} re-estimate the target priors by EM and test whether they differ from the source prior. BBSE~\cite{lipton2018bbse} inverts a confusion matrix estimated on labeled source data. GLA~\cite{zhu2023gla} fits the name bias of zero-shot CLIP as a per-class logit offset on labeled data. It adds the debiased logits to those of a fine-tuned model with equal weights, which is Bayes optimal if the two are independent given the label. Without labels, zero-shot predictions and pseudo-labels are rebalanced toward a uniform~\cite{wang2022debiaspl,liusie2023mitigating,qian2023inmap}, given~\cite{shin2024otter} or estimated marginal~\cite{dai2025freetta}. Wang \etal~\cite{wang2022solar} estimate such a marginal in partial-label learning~\cite{he2025tuning} and refine candidate-set labels toward it by Sinkhorn scaling. {\ours} asks the question of Saerens \etal without a source prior, namely whether the skew of the zero-shot marginal reflects the data. Since logits alone cannot separate name bias from true imbalance, it selects the raw or the debiased prior from the images. For fusion, GLA's equal weights rely on conditional independence, which fails because the targets of {\ours} come from zero-shot outputs. {\ours} therefore selects the fusion weight by closed-form LOO.

\section{Method}
\label{sec:method}
As illustrated in \cref{fig:overview}, for each task, {\ours} estimates the posterior of the label of every unlabeled image, as the targets of the classifier and in place of labels when the fusion weight is chosen. \Cref{sec:setup} fixes the setting and shows why posteriors are the right targets for the analytic classifier. The next three subsections answer questions (i) to (iii) of \cref{sec:intro} in the order of \cref{fig:overview}. Class-prior correction (\cref{sec:prior}) chooses the class proportions that the labels of a new task follow. Posterior label-distribution estimation (\cref{sec:em}) then refines each label distribution with the images of the task, and soft-label analytic learning and fusion (\cref{sec:head}) trains the classifier on these soft labels and selects its fusion weight after each task.

\subsection{Problem Formulation}
\label{sec:setup}

\paragraph{Setting.}
Following N2L~\cite{li2026n2l}, a model learns tasks $t=1,\dots,T$ in sequence, as when a shop or a camera-trap survey adds newly named products or species from unlabeled images. Task $t$ brings $k_t$ new classes $\mathcal{Y}_t$, disjoint from those of other tasks, with only their names and a set $\mathcal{D}_t$ of $n_t$ unlabeled images drawn from these classes. No label $y_i\in\mathcal{Y}_t$ is revealed, even in the first task, which holds a tenth of the classes in the zero-base setting ($T=10$) and half in the half-base setting ($T=6$). No earlier image or per-image feature is kept, and after task $t$ the model classifies test images over all $K_t$ seen classes $\mathcal{Y}_{1:t}=\mathcal{Y}_1\cup\dots\cup\mathcal{Y}_t$ without task identity. A frozen CLIP model with image encoder $f_{\mathrm{img}}$ and text encoder $f_{\mathrm{text}}$ gives the feature $x_i\in\mathbb{R}^d$ of image $i$ and the embedding $z_c=\nrm(f_{\mathrm{text}}(\mathrm{prompt}(c)))$ of class $c$, where $\nrm(v)=v/\|v\|$ and $\mathrm{prompt}(c)$ inserts the class name into the first prompt template. With CLIP's logit scale $s$ and the logits $\ell_{ic}=s\,\nrm(x_i)^\top z_c$, the raw zero-shot distribution of image $i$, row $\Praw_{i\cdot}$ of $\Praw\in\mathbb{R}^{n_t\times k_t}$, is \looseness=-1
\begin{equation}
\Praw_{ic}=\frac{\exp(\ell_{ic})}{\sum_{c'\in\mathcal{Y}_t}\exp(\ell_{ic'})},\qquad c\in\mathcal{Y}_t.
\label{eq:zs}
\end{equation}

\paragraph{Analytic Classifier.}
Like RanPAC~\cite{mcdonnell2023ranpac}, RAIL~\cite{xu2024rail} and N2L, {\ours} uses an analytic classifier, a ridge regression on random features $\phi(x)=\max(R^\top x,0)\in\mathbb{R}^D$ with a fixed random matrix $R\in\mathbb{R}^{d\times D}$. Each image $i$ of task $t$ enters with its features $\phi_i=\phi(x_i)$, a weight $m_i>0$ and a target $Q_i$ in the probability simplex $\Delta^{K_t-1}$, and after each task the classifier is updated in closed form~\cite{zhuang2022acil},
\begin{equation}
\begin{aligned}
A_t&=A_{t-1}+\textstyle\sum_{i\in\mathcal{D}_t}m_i\phi_i\phi_i^\top, \quad
W_t=(A_t+\lambda I_D)^{-1}C_t,\\
C_t&=\bigl[C_{t-1},\mathbf{0}\bigr]+\textstyle\sum_{i\in\mathcal{D}_t}m_i\phi_iQ_i^\top,
\end{aligned}
\label{eq:analytic}
\end{equation}
where $A_t\in\mathbb{R}^{D\times D}$ and $C_t\in\mathbb{R}^{D\times K_t}$ accumulate weighted moments ($A_0=0$, $C_0$ empty), $\mathbf{0}$ is a $D\times k_t$ zero block for the new classes, $\lambda>0$ is the ridge coefficient and $W_t\in\mathbb{R}^{D\times K_t}$ is the classifier, the ridge solution on all seen images although none is stored. Hard-label methods such as N2L set $Q_i$ to the one-hot vector $e_c$ of a pseudo-label $c$.

\paragraph{Posterior targets.}
The true label $y_i$ of each image is unknown, and the classifier $W_t^\star$ that the true labels would give is \cref{eq:analytic} with $Q_i=e_{y_i}$. Let $\mathcal{O}$ denote everything observed, the images and class names of all seen tasks, and let the weights $m_i$ be fixed. As $W_t$ is linear in its targets,
\begin{equation}
\mathbb{E}\bigl[W_t^\star\mid\mathcal{O}\bigr]=(A_t+\lambda I_D)^{-1}\textstyle\sum_{i}m_i\phi_i\,\mathbb{E}\bigl[e_{y_i}\mid\mathcal{O}\bigr]^\top,
\label{eq:posterior}
\end{equation}
where the sum runs over all seen images and $\mathbb{E}[e_{y_i}\mid\mathcal{O}]$ is the posterior distribution of $y_i$. Training on the posterior of each label therefore gives the conditional mean of the classifier that the true labels would give. Labeling becomes posterior estimation, and \cref{sec:prior,sec:em} estimate the posterior of each image of task $t$ as a soft label $q_i\in\Delta^{k_t-1}$.

\subsection{Class-Prior Correction}
\label{sec:prior}

The class proportions of a new task are unknown. Let $\bar p_t=\frac1{n_t}\sum_i\Praw_{i\cdot}\in\Delta^{k_t-1}$ be the class marginal of the zero-shot distribution and $u$ the uniform marginal with entries $1/k_t$. The marginal $\bar p_t$ departs from $u$ for two reasons that call for opposite actions. Name bias should be removed, since CLIP disfavors some class names, such as an aircraft variant named without its manufacturer, and gives these classes too little mass. True imbalance should be kept, since in datasets such as SUN the classes differ in size and the skew of $\bar p_t$ reflects the data. Both act on the logits as one offset per class, and the text alone cannot tell them apart. {\ours} therefore builds one candidate prior for each cause and selects between them from the images.

\paragraph{Two candidate priors.}
The raw candidate is $\Praw$ itself. The debiased candidate removes any class-level offset $\eta\in\mathbb{R}^{k_t}$ from the logits. Starting from $\eta=\mathbf 0$, Sinkhorn iterations~\cite{cuturi2013sinkhorn} alternate, for every $c\in\mathcal{Y}_t$,
\begin{equation}
\begin{aligned}
&P(\eta)_{ic}=\frac{\exp(\ell_{ic}+\eta_c)}{\sum_{c'\in\mathcal{Y}_t}\exp(\ell_{ic'}+\eta_{c'})},\\
&\eta_c \leftarrow \eta_c+\log\frac{n_t}{k_t}-\log\sum_{i\in\mathcal{D}_t}P(\eta)_{ic},
\end{aligned}
\label{eq:sinkhorn}
\end{equation}
until every class has mass $n_t/k_t$. As $P(\mathbf 0)=\Praw$, the iterations move from the raw distribution to the debiased distribution $\Pbal=P(\eta)$, the KL projection of $\Praw$ onto distributions with equal class masses~\cite{csiszar1975divergence}.

\paragraph{Prior selection rule.}
The prior selection rule tests whether the images reproduce the imbalance of $\bar p_t$ under a prior that carries none. It runs the estimator of \cref{sec:em} from the debiased prior without tempering, $q^{\mathrm{bal}}=\mathrm{EM}(\Pbal,1)$, which gives the likelihood its full weight, and takes the class marginal $\bar q_t=\frac1{n_t}\sum_iq^{\mathrm{bal}}_{i\cdot}$. Since $\Pbal$ has a uniform marginal and the estimator never re-estimates the prior of an image, any departure of $\bar q_t$ from $u$ comes from the images alone. The rule regresses this departure on that of the zero-shot marginal and selects the prior,
\begin{equation}
\rho_t=\frac{\langle\bar q_t-u,\,\bar p_t-u\rangle}{\|\bar p_t-u\|^2},\qquad
\Psel_t=\begin{cases}\Praw,&\rho_t>\frac12,\\ \Pbal,&\text{otherwise},\end{cases}
\label{eq:rho}
\end{equation}
with $\rho_t=0$ if $\bar p_t=u$. The threshold is not tuned, since $\rho_t>\frac12$ exactly when $\bar q_t$ is closer to $\bar p_t$ than to $u$. $\Psel_t$ fixes the class proportions of the task, but each of its rows rests on the zero-shot logits of a single image.

\subsection{Posterior Label-Distribution Estimation}
\label{sec:em}

Images of a class tend to lie close together, the cluster assumption~\cite{chapelle2006ssl}, which allows the images of a task to correct each other's label distributions. Fitting a standard mixture model to CLIP features faces two obstacles. Cosines between CLIP features mix class differences with within-class variation such as pose and background, and such a mixture would re-estimate the class proportions that \cref{sec:prior} has fixed. {\ours} therefore fits a mixture with fixed per-image priors on whitened features, whose statistics cover all seen classes without storing their images.

\paragraph{Whitening with accumulated statistics.}
{\ours} compares images on the unit sphere after the whitening map
\begin{equation}
\tilde x_i=\nrm\bigl(L^{-1}V^\top(x_i-\bar x)\bigr)\in\mathbb{S}^{d_p-1},
\label{eq:whiten}
\end{equation}
which centers $x_i$ at the mean $\bar x$ of all seen features, keeps its coordinates along the top $d_p$ principal directions $V\in\mathbb{R}^{d\times d_p}$ of their total covariance, and applies $L^{-1}$ to make the within-class spread equal in every direction, so that cosines on $\mathbb{S}^{d_p-1}$ become Mahalanobis cosines~\cite{goswami2023fecam,prabhu2024randumb}.
Here $L$ is the Cholesky factor of $V^\top\Sw V$, and $\Sw\in\mathbb{R}^{d\times d}$ is the within-class covariance after shrinkage toward a multiple of the identity with the oracle approximating shrinkage (OAS) coefficient~\cite{chen2010oas}, which keeps $L^{-1}$ stable.

With few images per task, all with uncertain labels, $\Sw$ is pooled over all seen classes.
Earlier images enter only through statistics accumulated under their frozen soft labels, the second moment $G$, the sum $g$ and the number $N$ of their features, and for each class $c$ the soft sum $h_c$, a column of $H$, and the soft count $\nu_c$. A run of EM merges the $n_t$ current images and their current soft labels $q$ into these statistics, and a prime marks the merged values, $G'=G+\sum_ix_ix_i^\top$, $g'=g+\sum_ix_i$ and $N'=N+n_t$, with $h'_c=\sum_iq_{ic}x_i$ and $\nu'_c=\sum_iq_{ic}$ for each new class $c\in\mathcal{Y}_t$ and $h'_c=h_c$ and $\nu'_c=\nu_c$ for earlier classes. They give $\bar x=g'/N'$, the directions $V$ as the top eigenvectors of $G'/N'-\bar x\bar x^\top$, and
\begin{equation}
\Sw\propto G'-\sum_{c:\,\nu'_c>0}\frac{h'_ch'^\top_c}{\nu'_c},
\label{eq:sw}
\end{equation}
the soft within-class scatter of all seen images, whose scale cancels in OAS and in \cref{eq:whiten}, so no earlier image is stored.

\paragraph{Fixed-prior vMF mixture.}
On this sphere each class $c\in\mathcal{Y}_t$ is a von Mises-Fisher (vMF) component with density proportional to $\exp(\kappa\mu_c^\top\tilde x)$, mean direction $\mu_c\in\mathbb{S}^{d_p-1}$ and a concentration $\kappa>0$ shared by all classes. The shared $\kappa$ cancels the normalizers, so the likelihood has no class-specific term, unlike a shared-covariance Gaussian whose term $-\frac12\|\mu_c\|^2$ favors classes with short means. For a prior $P\in\{\Pbal,\Psel_t\}$, the mixing weights of image $i$ are fixed to $P_{i\cdot}$. The class proportions chosen in \cref{sec:prior} enter every E-step and are never re-estimated. Starting from $q_{i\cdot}=P_{i\cdot}$, EM alternates between an M-step for the mean directions $\mu_c$ and the concentration $\kappa$, and an E-step for the soft labels,
\begin{equation}
\begin{aligned}
\mu_c&=\nrm\Bigl(\textstyle\sum_iq_{ic}\tilde x_i\Bigr),\qquad
\kappa=\frac{\bar r\,(d_p-\bar r^2)}{1-\bar r^2},\\
q_{ic}&=\frac{P_{ic}\exp\bigl(\kappa\mu_c^\top\tilde x_i/\tau\bigr)}{\sum_{c'\in\mathcal{Y}_t}P_{ic'}\exp\bigl(\kappa\mu_{c'}^\top\tilde x_i/\tau\bigr)},
\end{aligned}
\label{eq:em}
\end{equation}
where $\bar r=\frac1{n_t}\sum_c\|\sum_iq_{ic}\tilde x_i\|\in[0,1)$ is the mean resultant length and $\kappa$ is the approximation of~\cite{banerjee2005vmf} to its maximum-likelihood value. The E-step has the form of the assignment step of TransCLIP~\cite{zanella2024transclip} without its graph term. The temperature $\tau\ge1$ tempers the likelihood, so that small cosine gaps, scaled by the fitted $\kappa$, cannot override the text prior, and $\tau=1$ gives the untempered posterior. EM runs in two passes, the second recomputing $\Sw$, $L$ and $\tilde x_i$ from the first-pass posterior with $\bar x$ and $V$ fixed, since the prior alone gives only a coarse within-class estimate. This defines $q=\mathrm{EM}(P,\tau)$. The soft labels of task $t$ are $q=\mathrm{EM}(\Psel_t,\tau)$, after which $(x_i,q_i)$ are added to $(G,g,N,H,\nu)$ and never revised.

\subsection{Soft-Label Analytic Learning and Fusion}
\label{sec:head}

The soft labels of a task are used twice before they are discarded. They first train the analytic classifier and then, in place of labels, select the weight that fuses it with zero-shot CLIP in predictions over all seen classes.

\paragraph{Soft-label analytic learning.}
Following \cref{eq:posterior}, the target of image $i$ is its soft label padded with zeros on the old classes, $Q_i=(\mathbf 0_{K_{t-1}}^\top,q_i^\top)^\top$. For any output $o\in\mathbb{R}^{K_t}$, $\mathbb{E}_{y\sim Q_i}\|o-e_y\|^2=\|o-Q_i\|^2+1-\|Q_i\|^2$, so the squared error of \cref{eq:analytic} fits the label distribution rather than its argmax. To keep large classes from dominating, the weights balance classes by their soft counts $\hat n_c=\sum_iq_{ic}$,
\begin{equation}
m_i=\sum_{c\in\mathcal{Y}_t}q_{ic}\,f_c,\qquad
f_c=\frac{n_t}{\Keff\max(\hat n_c,1)},
\label{eq:weights}
\end{equation}
where $\Keff=\max(|\{c:\hat n_c\ge1\}|,1)$ is the number of classes with at least unit mass, and $f_c$ equals the class-balancing factor of N2L for one-hot labels.

\paragraph{Fusion weight selection.}
At test time, {\ours} predicts the class in $\mathcal{Y}_{1:t}$ with the largest fused score
\begin{equation}
F_t(x)=(1-\omega_t)\,\pzsbar(x)+\omega_t\,W_t^\top\phi(x),
\label{eq:fusion}
\end{equation}
where $\pzsbar(x)\in\Delta^{K_t-1}$ is the zero-shot distribution over all seen classes with each class embedding averaged over all prompt templates, and $\omega_t\in[0,1]$ weights the classifier.
The right weight changes across datasets and tasks. Choosing it requires predictions that, like test predictions, come from a classifier that has not seen the image.
The fitted output $r_i=W_t^\top\phi_i$ does not qualify, since each image has pulled $W_t$ toward its own target.
{\ours} instead uses the closed-form LOO output of ridge regression~\cite{allen1974press,hastie2009esl}, \looseness=-1
\begin{equation}
\tilde r_i=\frac{r_i-h_iQ_i}{1-h_i},\qquad h_i=m_i\,\phi_i^\top(A_t+\lambda I_D)^{-1}\phi_i,
\label{eq:loo}
\end{equation}
where $h_i\in[0,1)$ is the leverage of image $i$ and $\tilde r_i\in\mathbb{R}^{K_t}$ is the output at $x_i$ of the classifier re-solved with $m_i=0$.
After training on task $t$, the soft labels score the fused LOO decisions in place of the labels,
\begin{equation}
\begin{aligned}
\hat c_i(\omega)&=\argmax_{c\in\mathcal{Y}_{1:t}}\bigl[(1-\omega)\,\pzs_c(x_i)+\omega\,\tilde r_{ic}\bigr],\\
\omega_t&=\argmax_{\omega\in\Omega}\ \textstyle\sum_{i\in\mathcal{D}_t}Q_{i,\hat c_i(\omega)},
\end{aligned}
\label{eq:omega}
\end{equation}
where $\pzs_c(x_i)$ is the softmax of $\ell_{ic}$ over all seen classes $c\in\mathcal{Y}_{1:t}$ and $\Omega=\{0,0.01,\dots,1\}$, with ties resolved by their median. Each decision earns the probability that its soft label gives to the chosen class, so the criterion counts the correct decisions expected under the soft labels, and a decision for an old class earns nothing. Across tasks, {\ours} stores only $(G,g,N,H,\nu)$ and $(A_t,C_t)$, whose size does not grow with the number of seen images, and \cref{alg:task} in Supp.~\ref{supp:impl} lists every step.

\paragraph{Theoretical analysis.}
Supp.~\ref{sec:theory} justifies both uses of the soft labels, the first in squared error and the second in expected correct decisions.
As targets, the posteriors minimize the mean-square gap to the true-label prediction, and any other targets add at most a multiple of their squared distance to them.
Their argmax adds exactly the squared change it causes in the prediction.
{\ours} therefore trains on soft labels, and \cref{sec:prior,sec:em} aim to bring them close to the posteriors.
In place of labels, the soft labels make the criterion of \cref{eq:omega} the expected number of correct LOO decisions up to their error, which bounds what $\omega_t$ can lose.

\section{Experiments}
\label{sec:exp}

\begin{table*}[t]
\caption{Comparison with exemplar-free methods on six datasets. $\bar A$ and $A_T$ in \%, and Mean averages the six settings of each block.}
\label{tab:main}
\centering\footnotesize
\setlength{\tabcolsep}{5pt}
\begin{tabular}{@{}llcccccccccccccc@{}}
\toprule
\multirow{3}{*}[-1.25ex]{Method} & \multirow{3}{*}[-1.25ex]{Venue} & \multicolumn{4}{c}{Aircraft} & \multicolumn{4}{c}{Cars} & \multicolumn{4}{c}{CIFAR100} & \multicolumn{2}{c}{\multirow{2}{*}[-0.7ex]{Mean}}\\
\cmidrule(lr){3-6}\cmidrule(lr){7-10}\cmidrule(lr){11-14}
 & & \multicolumn{2}{c}{B0Inc10} & \multicolumn{2}{c}{B50Inc10} & \multicolumn{2}{c}{B0Inc10} & \multicolumn{2}{c}{B50Inc10} & \multicolumn{2}{c}{B0Inc10} & \multicolumn{2}{c}{B50Inc10} & \multicolumn{2}{c}{}\\
\cmidrule(lr){3-4}\cmidrule(lr){5-6}\cmidrule(lr){7-8}\cmidrule(lr){9-10}\cmidrule(lr){11-12}\cmidrule(lr){13-14}\cmidrule(l){15-16}
 & & $\bar A$ & $A_T$ & $\bar A$ & $A_T$ & $\bar A$ & $A_T$ & $\bar A$ & $A_T$ & $\bar A$ & $A_T$ & $\bar A$ & $A_T$ & $\bar A$ & $A_T$\\
\midrule
ZS-CLIP & ICML'21 & 26.61 & 17.16 & 21.66 & 17.16 & 82.90 & 76.73 & 78.74 & 76.73 & 81.81 & 71.38 & 76.49 & 71.38 & 61.37 & 55.09\\
CODA-Prompt & CVPR'23 & 26.56 & 17.79 & 18.11 & 12.28 & 79.69 & 69.03 & 67.81 & 56.59 & 79.36 & 67.74 & 66.94 & 54.64 & 56.41 & 46.35\\
MoE-Adapter & CVPR'24 & 26.73 & 17.92 & 22.19 & 17.52 & 83.16 & 77.02 & 79.21 & 77.72 & 81.86 & 71.53 & 76.57 & 71.49 & 61.62 & 55.53\\
RAPF & ECCV'24 & 29.07 & 19.52 & 20.83 & 18.46 & 83.77 & 73.26 & 75.73 & 71.56 & 84.47 & 76.08 & 77.30 & 74.33 & 61.86 & 55.54\\
RAIL & NeurIPS'24 & 36.23 & 33.59 & 23.75 & 25.99 & 88.64 & 84.68 & 82.72 & 82.13 & 87.34 & 80.37 & 81.44 & 78.89 & 66.69 & 64.28\\
ENGINE & ICCV'25 & 34.77 & 25.41 & 26.94 & 23.56 & 86.90 & 78.76 & 82.67 & 79.93 & 85.15 & 77.11 & 79.89 & 76.15 & 66.05 & 60.15\\
N2L & ICLR'26 & \underline{43.73} & \underline{40.21} & \underline{29.69} & \underline{32.42} & \underline{92.38} & \underline{87.50} & \underline{86.42} & \underline{85.45} & \underline{87.80} & \underline{81.13} & \underline{82.92} & \underline{80.30} & \underline{70.49} & \underline{67.84}\\
{\ours} & Ours & \textbf{51.35} & \textbf{49.70} & \textbf{36.75} & \textbf{39.80} & \textbf{92.73} & \textbf{89.40} & \textbf{88.53} & \textbf{87.95} & \textbf{88.82} & \textbf{82.71} & \textbf{83.87} & \textbf{81.49} & \textbf{73.68} & \textbf{71.84}\\
\midrule
\rowcolor{labelrow}\labelrowfirst{N2L+Label} & ICLR'26 & 66.38 & 56.31 & 59.18 & 56.31 & 93.57 & 89.15 & 90.92 & 89.14 & 88.52 & 81.92 & 85.23 & 81.92 & 80.63 & \labelrowlast{75.79}\\
\rowcolor{labelrow}\labelrowfirst{{\ours}+Label} & Ours & 71.39 & 61.78 & 65.43 & 61.78 & 94.27 & 91.21 & 92.17 & 91.12 & 89.68 & 83.55 & 86.50 & 83.55 & 83.24 & \labelrowlast{78.83}\\
\midrule[\heavyrulewidth]
\multirow{3}{*}[-1.25ex]{Method} & \multirow{3}{*}[-1.25ex]{Venue} & \multicolumn{4}{c}{CUB} & \multicolumn{4}{c}{ObjectNet} & \multicolumn{4}{c}{UCF} & \multicolumn{2}{c}{\multirow{2}{*}[-0.7ex]{Mean}}\\
\cmidrule(lr){3-6}\cmidrule(lr){7-10}\cmidrule(lr){11-14}
 & & \multicolumn{2}{c}{B0Inc20} & \multicolumn{2}{c}{B100Inc20} & \multicolumn{2}{c}{B0Inc20} & \multicolumn{2}{c}{B100Inc20} & \multicolumn{2}{c}{B0Inc10} & \multicolumn{2}{c}{B50Inc10} & \multicolumn{2}{c}{}\\
\cmidrule(lr){3-4}\cmidrule(lr){5-6}\cmidrule(lr){7-8}\cmidrule(lr){9-10}\cmidrule(lr){11-12}\cmidrule(lr){13-14}\cmidrule(l){15-16}
 & & $\bar A$ & $A_T$ & $\bar A$ & $A_T$ & $\bar A$ & $A_T$ & $\bar A$ & $A_T$ & $\bar A$ & $A_T$ & $\bar A$ & $A_T$ & $\bar A$ & $A_T$\\
\midrule
ZS-CLIP & ICML'21 & 75.47 & 63.72 & 69.06 & 63.72 & 38.43 & 26.43 & 31.12 & 26.43 & 75.88 & 67.79 & 71.68 & 67.79 & 60.27 & 52.65\\
CODA-Prompt & CVPR'23 & 67.40 & 51.91 & 53.24 & 37.64 & 38.25 & 26.23 & 26.22 & 19.22 & 81.61 & 75.18 & 67.06 & 58.87 & 55.63 & 44.84\\
MoE-Adapter & CVPR'24 & 73.69 & 63.07 & 68.29 & 63.09 & 40.14 & 28.70 & 32.45 & 28.53 & 75.35 & 67.36 & 72.28 & 69.07 & 60.37 & 53.30\\
RAPF & ECCV'24 & 72.08 & 58.14 & 62.27 & 56.02 & 39.16 & 25.88 & 28.93 & 24.70 & 85.99 & 81.10 & 75.70 & 74.31 & 60.69 & 53.36\\
RAIL & NeurIPS'24 & 81.64 & 73.93 & 73.08 & 70.91 & 39.80 & 35.13 & 31.21 & 30.19 & 90.18 & 89.90 & 81.05 & 84.27 & 66.16 & 64.06\\
ENGINE & ICCV'25 & 77.06 & 65.07 & 70.85 & 64.92 & 44.57 & 31.24 & 34.62 & 29.96 & 87.85 & 84.46 & 82.30 & 81.04 & 66.21 & 59.45\\
N2L & ICLR'26 & \underline{83.41} & \underline{76.48} & \underline{75.16} & \underline{73.40} & \underline{49.31} & \underline{41.59} & \underline{41.42} & \textbf{38.65} & \underline{95.00} & \underline{93.29} & \underline{86.41} & \underline{87.87} & \underline{71.79} & \underline{68.55}\\
{\ours} & Ours & \textbf{86.71} & \textbf{80.38} & \textbf{78.76} & \textbf{77.21} & \textbf{53.06} & \textbf{41.62} & \textbf{41.88} & \underline{37.78} & \textbf{95.92} & \textbf{94.72} & \textbf{89.35} & \textbf{90.40} & \textbf{74.28} & \textbf{70.35}\\
\midrule
\rowcolor{labelrow}\labelrowfirst{N2L+Label} & ICLR'26 & 86.43 & 79.05 & 82.29 & 79.06 & 53.18 & 45.27 & 49.71 & 45.26 & 98.74 & 97.75 & 98.37 & 97.75 & 78.12 & \labelrowlast{74.02}\\
\rowcolor{labelrow}\labelrowfirst{{\ours}+Label} & Ours & 87.43 & 81.23 & 83.89 & 81.24 & 58.89 & 46.19 & 51.67 & 46.15 & 98.67 & 97.40 & 98.22 & 97.36 & 79.80 & \labelrowlast{74.93}\\
\bottomrule
\end{tabular}
\end{table*}

We first describe the setup (\cref{sec:exp-setup}), then compare {\ours} with exemplar-free methods on six datasets (\cref{sec:exp-main}) and ablate its three modules (\cref{sec:exp-ablation}). The supplement compares {\ours} on Food, ImageNet-R and SUN and with exemplar-based methods (Supp.~\ref{supp:comparisons}), tests its robustness to $\tau$, $\lambda$, the CLIP backbone (Supp.~\ref{supp:robust}) and noisy task boundaries (Supp.~\ref{supp:noisy}), gives N2L two components of {\ours} (Supp.~\ref{supp:transfer}) and {\ours} the soft labels of transductive methods (Supp.~\ref{supp:transductive}), and reports pseudo-label quality, per-task accuracy, and training and inference cost (Supp.~\ref{supp:pl-quality} to \ref{supp:cost}).

\subsection{Experimental Setup}
\label{sec:exp-setup}

\paragraph{Datasets.}
We use the nine datasets with the class subsets and train/test splits adopted from N2L~\cite{li2026n2l}, PROOF~\cite{zhou2025proof} and ENGINE~\cite{zhou2025engine}. They cover fine-grained categories in Aircraft~\cite{maji2013aircraft}, Cars~\cite{krause2013cars}, CUB~\cite{wah2011cub} and Food~\cite{bossard2014food}, generic objects in CIFAR100~\cite{krizhevsky2009cifar}, distribution shift in ImageNet-R~\cite{hendrycks2021imagenetr} and ObjectNet~\cite{barbu2019objectnet}, human actions in UCF~\cite{soomro2012ucf101} and scenes in SUN~\cite{xiao2010sun}. SUN has 300 classes, CUB, ObjectNet and ImageNet-R have 200, and the others have 100. B$b$Inc$k$ denotes a first task of $b$ classes followed by tasks of $k$ classes, with $b=0$ meaning that the first task also has $k$ classes. The \emph{zero-base setting} splits all classes evenly into 10 tasks (B0Inc10, B0Inc20 or B0Inc30 for 100, 200 or 300 classes). The \emph{half-base setting} puts half of the classes in the first task and splits the other half evenly into 5 tasks (B50Inc10, B100Inc20 or B150Inc30). Each task, including the first, provides only the names of its new classes and their unlabeled training images.

\vspace{-3mm}
\paragraph{Evaluation Metrics.}
Following N2L, after each task $t$ the model classifies the test images of all classes seen so far without task identity. Its accuracy is $A_t=\frac{1}{|\mathcal{T}_{1:t}|}\sum_{(x,y)\in\mathcal{T}_{1:t}} \mathbb{I}[\hat y_t(x)=y]$, where $\mathcal{T}_{1:t}$ is the set of test images $x$ with labels $y$ in the seen classes $\mathcal{Y}_{1:t}$, $\hat y_t(x)\in\mathcal{Y}_{1:t}$ is the class predicted after task $t$, and the indicator $\mathbb{I}[\cdot]$ is 1 if its argument holds and 0 otherwise. We report the last accuracy $A_T$ after the final task $T$ and the average incremental accuracy $\bar A=\frac{1}{T}\sum_{t=1}^{T}A_t$, both in percent.

\vspace{-3mm}
\paragraph{Comparison Methods.}
We compare with the exemplar-free methods evaluated by N2L. ZS-CLIP~\cite{radford2021clip} classifies by class names without training. CODA-Prompt~\cite{smith2023coda}, MoE-Adapter~\cite{yu2024moe}, RAPF~\cite{huang2024rapf}, ENGINE~\cite{zhou2025engine} and RAIL~\cite{xu2024rail} are designed for labeled tasks, so N2L trains them on the argmax of the zero-shot prediction of each image as its pseudo-label. CODA-Prompt tunes prompts of the image encoder, MoE-Adapter, RAPF and ENGINE (without its re-ranking) adapt CLIP by gradient descent, and RAIL fits an analytic classifier. N2L~\cite{li2026n2l} starts from the same pseudo-labels, refines them in three iterations with a ridge classifier on dimension-reduced features, and weights each image by its class frequency and confidence. We also compare with the exemplar-based PROOF~\cite{zhou2025proof} and CLAP4CLIP~\cite{jha2024clap4clip}, which store 20 images per class of earlier tasks (Supp.~\ref{supp:exemplar}). N2L+Label and {\ours}+Label replace the pseudo-labels of N2L and the soft labels of {\ours} by ground-truth labels, and serve as reference upper bounds that are not ranked.

\vspace{-3mm}
\paragraph{Implementation Details.}
{\ours} uses the frozen CLIP ViT-B/16 of N2L, pre-trained on LAION-400M~\cite{schuhmann2021laion}, and its random features with $D=5000$. As in N2L, class names are embedded with the first prompt template of each dataset during training and with the ensemble of all its templates at test time. All datasets share one hyperparameter setting, $\tau=4$, $\lambda=10$ and $d_p=128$, and after training on each task, $\omega_t$ is selected from the grid $\{0,0.01,\dots,1\}$ by the closed-form LOO criterion of \cref{eq:omega}. Images are encoded in batches of 64. Each result is the mean of three runs with seeds 1993, 1994 and 1995 on one NVIDIA A100 GPU with PyTorch. The best and second-best results in each column are marked in \textbf{bold} and \underline{underlined}, respectively. We will release the code upon the manuscript acceptance.

\subsection{Main Results}
\label{sec:exp-main}

\Cref{tab:main} covers the six datasets on which N2L evaluates all baselines. Among the methods trained without labels, {\ours} has the best $\bar A$ on all 12 settings and the best $A_T$ on 11 of them. N2L is the strongest compared method throughout, and in the Mean columns {\ours} improves on it by 3.19\% and 2.49\% in $\bar A$ on the two blocks and by 4.00\% and 1.80\% in $A_T$, the latter held down by ObjectNet, with $\bar A$ gains of 2.83\% and 2.85\% on average in the zero-base and half-base settings. The gains over N2L are largest on Aircraft, 7.62\% and 7.06\% in $\bar A$ and 9.49\% and 7.38\% in $A_T$ on B0Inc10 and B50Inc10. On this dataset zero-shot CLIP is the weakest, with an $A_T$ of 17.16\%, and ground-truth labels raise the $\bar A$ of N2L the most, by 22.65\% and 29.49\%. The pseudo-labels of N2L are least reliable here. This is the case that motivates {\ours} in \cref{sec:intro}, since the zero-shot distribution keeps plausible classes that the argmax discards. {\ours} falls behind only in $A_T$ on ObjectNet B100Inc20, 0.87\% below N2L, and on ObjectNet B0Inc20 it gains 3.75\% in $\bar A$ but 0.03\% in $A_T$. ObjectNet has the second weakest zero-shot CLIP and varies rotation, background and viewpoint by design, so both the zero-shot prior and the feature clusters used by EM are weak, and the early gain of {\ours} fades by the last task.

\begin{figure}[!t]
    \centering
	\begin{overpic}[width=8.3cm]{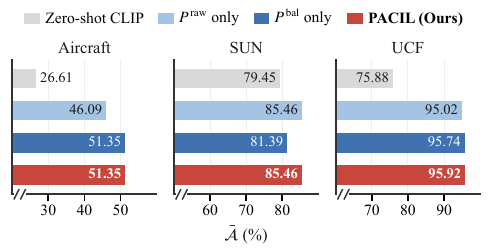}
    \end{overpic}
    \caption{Ablation of class-prior correction.
$\bar A$ (\%) on the zero-base settings of Aircraft, SUN and UCF with $\Praw$ or $\Pbal$ fixed on every task and with the prior selection rule of {\ours}.}
 \label{fig:ablation_prior}
\end{figure}

The +Label rows measure the cost of the missing labels. {\ours}+Label is above {\ours} by 5.29\% and 9.79\% in $\bar A$ on average in the zero-base and half-base settings, and by 3.81\% and 7.76\% in $A_T$. {\ours}+Label reaches nearly the same $A_T$ in both settings, whereas {\ours} is lower in the half-base setting on all six datasets. Its first task must then be labeled as a 50-way or 100-way problem instead of a 10-way or 20-way one, and labels that enter the classifier are never corrected. The gap is largest on Aircraft, 20.04\% and 28.68\% in $\bar A$, and below 1\% in both metrics on CIFAR100 B0Inc10 and CUB B0Inc20, where {\ours} is also above N2L+Label. {\ours}+Label is in turn above N2L+Label in both metrics on 10 of the 12 settings, all except UCF.

\subsection{Ablation Study}
\label{sec:exp-ablation}
\paragraph{Class-Prior Correction.}
We replace the prior selection rule of \cref{eq:rho} by $\Praw$ or $\Pbal$ on every task ($\Praw$ only, $\Pbal$ only). Each fails on a different dataset in \cref{fig:ablation_prior}.
$\Praw$ only loses 5.26\% in $\bar A$ on Aircraft B0Inc10, where the classes are balanced and the skew of the zero-shot marginal is name bias. $\Pbal$ only loses 4.07\% on SUN B0Inc30, where the classes differ in size and debiasing erases a real imbalance. The prior selection rule matches the better fixed prior on both, and on UCF B0Inc10 it switches priors across tasks and exceeds both. Its gain over the better fixed prior is small, however, none on Aircraft and SUN and 0.18\% on UCF, so the rule mainly guards against a wrong fixed prior.

\begin{figure}[!t]
    \centering
	\begin{overpic}[width=8.3cm]{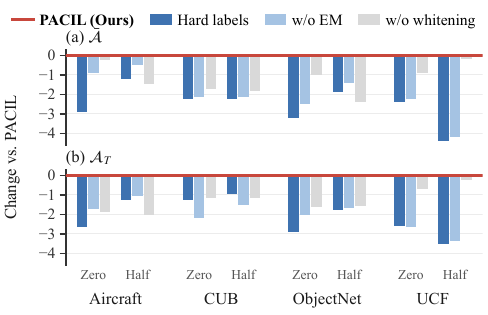}
    \end{overpic}
    \caption{Ablation of posterior label-distribution estimation.
Change in (a)~$\bar A$ and (b)~$A_T$ (\%) relative to {\ours} when hard pseudo-labels replace the label posteriors (Hard labels), the final EM is skipped (w/o EM) or run with $L=I$ (w/o whitening), on the zero-base (Zero) and half-base (Half) settings of four datasets.}
 \label{fig:ablation_posterior}
\end{figure}

\paragraph{Posterior Label-Distribution Estimation.}
Keeping the prior selection rule, we change only how $\Psel_t$ becomes the soft labels. \emph{Hard labels} replaces the label posteriors with hard pseudo-labels, \emph{w/o EM} trains directly on $q_i=\Psel_{i\cdot}$, and \emph{w/o whitening} runs EM with $L=I$ in \cref{eq:whiten}. All three are below {\ours} in both metrics on the eight settings of \cref{fig:ablation_posterior}. Removing EM costs up to 4.18\% in $\bar A$ on UCF B50Inc10, in line with the cluster assumption that the images of a task correct each other's distributions (\cref{sec:em}). Without whitening, EM loses up to 2.41\% on ObjectNet B100Inc20, whose images vary in pose and background, but about 0.2\% on UCF B50Inc10 and Aircraft B0Inc10. Taking the hard labels lowers $\bar A$ further on all eight settings, most on Aircraft B0Inc10 with 1.99\%, where the distribution keeps plausible classes that the weak zero-shot argmax misses. Keeping the whole distribution helps before EM is run, and EM then brings it closer to the posterior of each label. In $A_T$, the argmax is better than w/o EM on both CUB settings, by 0.95\% and 0.54\%.

\begin{figure}[!t]
    \centering
	\begin{overpic}[width=8.3cm]{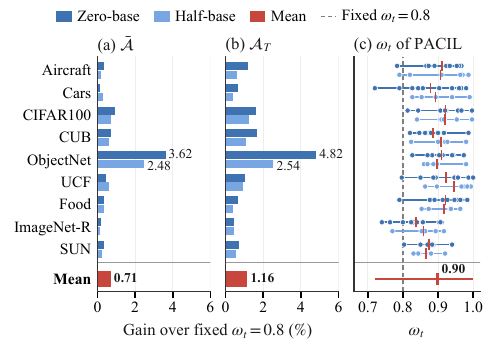}
    \end{overpic}
    \caption{Ablation of fusion weight selection.
(a, b)~Gain of {\ours} over the fixed $\omega_t=0.8$ of N2L in $\bar A$ and $A_T$ (\%) on all 18 settings.
(c)~$\omega_t$ selected on each task (dots, mean over three runs) and its mean per setting (ticks).
Bottom rows give the means over all settings, and in (c) the line spans the range of $\omega_t$.}
 \label{fig:ablation_fusion_weight}
\end{figure}

\paragraph{Fusion Weight Selection.}
We fix $\omega_t=0.8$ on every task as N2L does, which leaves the classifier unchanged. Selecting $\omega_t$ by \cref{eq:omega} is better on every setting, by 0.71\% in $\bar A$ and 1.16\% in $A_T$ on average (\cref{fig:ablation_fusion_weight}). Across the 144 tasks, the selected $\omega_t$ ranges from 0.72 to 1.00 with a mean of 0.90, and it is above the fixed 0.8 of N2L on 134 of them (\cref{fig:ablation_fusion_weight}c). The gain is small on ImageNet-R, 0.16\% and 0.14\% in $\bar A$, whose mean $\omega_t$ of about 0.84 and 0.86 is nearest 0.8. It is largest on ObjectNet, 3.62\% and 2.48\% in $\bar A$, though its mean $\omega_t$ is near 0.90, so ObjectNet is more sensitive to the weight than in need of a different one.

\section{Conclusion}
\label{sec:conclusion}

When class-incremental learning has only class names and unlabeled images, each image should carry its label posterior as its training target. An analytic classifier is linear in its targets, so training it on posteriors recovers the true-label classifier in expectation. {\ours} estimates them with a prior selection rule and a fixed-prior vMF mixture on whitened features, then uses the soft labels as targets and to select the fusion weight by closed-form LOO. In our analysis, no targets beat the posteriors in mean square, and rounding them to hard labels adds error only through ambiguous images. The error of the soft labels bounds the loss of the selected weight. {\ours} leads methods trained without labels in average incremental accuracy, most clearly where zero-shot CLIP is weakest, and its lead over N2L persists across six CLIP backbones. Its corrections are limited by how well CLIP relates class names to images, and a large gap to labeled training remains. Language descriptions or queried labels~\cite{gan2025semiclip} could narrow this gap, and future work should also admit unlabeled images from unknown classes.

{
    \small
    \bibliographystyle{ieeenat_fullname_unsrt}
    \bibliography{refs_cil}
}

\clearpage
\maketitlesupplementary

\setcounter{section}{0}
\renewcommand{\thesection}{\Alph{section}}
\renewcommand{\theHsection}{supp.\Alph{section}}
\setcounter{table}{0}
\renewcommand{\thetable}{S\arabic{table}}
\renewcommand{\theHtable}{supp.\arabic{table}}
\setcounter{figure}{0}
\renewcommand{\thefigure}{S\arabic{figure}}
\renewcommand{\theHfigure}{supp.\arabic{figure}}
\setcounter{equation}{0}
\renewcommand{\theequation}{S\arabic{equation}}
\renewcommand{\theHequation}{supp.\arabic{equation}}
\setcounter{theorem}{0}
\renewcommand{\thetheorem}{S\arabic{theorem}}
\providecommand{\theHtheorem}{}
\renewcommand{\theHtheorem}{supp.theorem.\arabic{theorem}}
\setcounter{lemma}{0}
\renewcommand{\thelemma}{S\arabic{lemma}}
\providecommand{\theHlemma}{}
\renewcommand{\theHlemma}{supp.lemma.\arabic{lemma}}

\renewcommand{\dbltopfraction}{0.9}
\renewcommand{\topfraction}{0.9}
\renewcommand{\textfraction}{0.07}
\renewcommand{\dblfloatpagefraction}{0.85}
\renewcommand{\floatpagefraction}{0.85}

\section{Theoretical Analysis}
\label{sec:theory}

Without labels, {\ours} must decide what its analytic classifier is trained on and how much its prediction trusts this classifier.
\Cref{thm:soft} answers both questions for the prediction that {\ours} deploys after each task, and \cref{lem:loo} shows that the leave-one-out (LOO) outputs that select the fusion weight come from classifiers that have not seen the image, just as test predictions do.
\Cref{supp:theory-statement} formulates the problem and states both, \cref{supp:theory-discussion} discusses what they imply for {\ours}, and \cref{supp:theory-proofs} proves them.

\subsection{Problem Formulation and Main Theorem}
\label{supp:theory-statement}

After task $t$, every seen image $j$ has its features $\phi_j$, its weight $m_j>0$ and its soft label $Q_j$, padded with zeros on the other classes of $\mathcal{Y}_{1:t}$ and frozen after its task.
For targets $\Theta=(\Theta_j)_j$, one vector $\Theta_j\in\mathbb{R}^{K_t}$ per seen image, the classifier fitted to them is
\begin{equation}
W(\Theta)=(A_t+\lambda I_D)^{-1}\sum_jm_j\phi_j\Theta_j^\top,
\end{equation}
and the classifier of \cref{eq:analytic} is $W_t=W(Q)$ by Step~1 of the proof.
Given everything observed, the only unknowns are the true labels.
For any joint law of images and labels, we therefore condition on the $\sigma$-algebra $\mathcal{F}_t$ generated by the observations $\mathcal{O}$ after task $t$ and by $R$, and write $\mathbb{E}_t$ for the expectation given $\mathcal{F}_t$.
Every quantity that {\ours} computes, such as $Q$, the weights, the zero-shot scores and $\omega_t$, is then fixed, while the label of image $j$ has the posterior
\begin{equation}
\bar e_j=\mathbb{E}_t[e_{y_j}],\qquad\bar e_{jc}=\Pr(y_j=c\mid\mathcal{F}_t).
\end{equation}
With the classifier fitted to $\Theta$, \cref{eq:fusion} predicts
\begin{equation}
F_\Theta(x)=(1-\omega_t)\,\pzsbar(x)+\omega_t\,W(\Theta)^\top\phi(x),
\end{equation}
so {\ours} predicts with $F_t=F_Q$, and the true labels $e_y=(e_{y_j})_j$ would give $F_{e_y}$ with the same $m_j$ and $\omega_t$.
We compare targets by their risk $R_x(\Theta)$ at an input $x$, and write $s_t(x)$ for the leverage of $x$ and $\sigma_t^2(x)$ for the conditional variance of the output that the true labels would give, which no choice of targets affects,
\begin{equation}
\begin{aligned}
R_x(\Theta)&=\mathbb{E}_t\|F_\Theta(x)-F_{e_y}(x)\|^2,\\
s_t(x)&=\phi(x)^\top(A_t+\lambda I_D)^{-1}\phi(x),\\
\sigma_t^2(x)&=\mathbb{E}_t\|(W(e_y)-W(\bar e))^\top\phi(x)\|^2.
\end{aligned}
\end{equation}
For the fusion weight, let $\hat c_i(\omega)$ be the decision of \cref{eq:omega} for a current image $i$, so that $\omega_t$ maximizes the criterion $J(\omega)$ and $N_{\mathrm{cor}}(\omega)$ counts the correct decisions, where
\begin{equation}
\begin{aligned}
J(\omega)&=\sum_{i\in\mathcal{D}_t}Q_{i,\hat c_i(\omega)},\\
N_{\mathrm{cor}}(\omega)&=\sum_{i\in\mathcal{D}_t}\mathbb{I}[\hat c_i(\omega)=y_i].
\end{aligned}
\end{equation}
Let $\omega^\star\in\Omega$ maximize $\mathbb{E}_t[N_{\mathrm{cor}}(\omega)]$, and let $\mathcal{S}$ hold the current images that $\omega_t$ and $\omega^\star$ decide differently.

\begin{theorem}[Keep the posterior, not the argmax]
\label{thm:soft}
\textup{\textbf{(a)}}~The posteriors give the conditional mean, $\mathbb{E}_t[F_{e_y}(x)]=F_{\bar e}(x)$, and for every input $x$ and all $\mathcal{F}_t$-measurable targets $\Theta$,
\begin{equation}
\begin{aligned}
R_x(\Theta)&=\omega_t^2\bigl(\|(W(\Theta)-W(\bar e))^\top\phi(x)\|^2+\sigma_t^2(x)\bigr)\\
&\le\omega_t^2\Bigl(\sigma_t^2(x)+s_t(x)\sum_jm_j\|\Theta_j-\bar e_j\|^2\Bigr),
\end{aligned}
\label{eq:thm-train}
\end{equation}
with $s_t(x)\le\|\phi(x)\|^2/\lambda$, so the posteriors attain the smallest risk, $R_x(\bar e)=\omega_t^2\sigma_t^2(x)$.
\textup{\textbf{(b)}}~Let $\hat Q_j=e_{c_j}$ with $c_j\in\argmax_cQ_{jc}$ be the hard label of image $j$, and let
\begin{equation}
\begin{aligned}
\delta_x&=(W(\hat Q)-W(Q))^\top\phi(x),\\
\epsilon_x&=\|(W(Q)-W(\bar e))^\top\phi(x)\|
\end{aligned}
\end{equation}
be the change of the output that the argmax causes, which needs no label, and the error of the soft labels at $x$.
Then
\begin{equation}
\begin{gathered}
\bigl|R_x(\hat Q)-R_x(Q)-\omega_t^2\|\delta_x\|^2\bigr|\le2\omega_t^2\|\delta_x\|\,\epsilon_x,\\
\epsilon_x^2\le s_t(x)\sum_jm_j\|Q_j-\bar e_j\|^2,
\end{gathered}
\label{eq:thm-argmax}
\end{equation}
so $R_x(\hat Q)-R_x(Q)=\omega_t^2\|\delta_x\|^2$ when $Q=\bar e$.
\textup{\textbf{(c)}}~For every $\omega\in\Omega$, the criterion splits as
\begin{equation}
J(\omega)=\mathbb{E}_t[N_{\mathrm{cor}}(\omega)]+\sum_{i\in\mathcal{D}_t}(Q_i-\bar e_i)_{\hat c_i(\omega)},
\label{eq:thm-crit}
\end{equation}
and the selected weight satisfies
\begin{equation}
0\le\mathbb{E}_t\bigl[N_{\mathrm{cor}}(\omega^\star)-N_{\mathrm{cor}}(\omega_t)\bigr]\le\sum_{i\in\mathcal{S}}\|Q_i-\bar e_i\|_1.
\label{eq:thm-select}
\end{equation}
\end{theorem}

\begin{lemma}[Closed-form LOO]
\label{lem:loo}
For $\lambda>0$, the leverage $h_i$ of \cref{eq:loo} lies in $[0,1)$, and $\tilde r_i$ is the output at $x_i$ of the classifier of \cref{eq:analytic} re-solved with $m_i=0$, all other seen images keeping their weights and targets.
\end{lemma}

\subsection{Discussion}
\label{supp:theory-discussion}

By (a), the posteriors are the targets that bring the deployed prediction closest, in mean square, to the one that the true labels would give, and other targets add at most $\omega_t^2s_t(x)$ times their weighted squared distance to the posteriors.
The sum runs over all seen images, and a soft label is frozen after its task while its image is discarded, so an error made on one task stays in the bound for every later task.
{\ours} therefore estimates the posteriors, with the prior selection rule of \cref{sec:prior} and the EM of \cref{sec:em}, before it freezes the soft labels for all later tasks.
By (b), if the soft labels are the posteriors, replacing them by their argmax raises the risk by exactly $\omega_t^2\|\delta_x\|^2$, the squared change of the prediction.
If they are estimates of the posteriors, the argmax raises the risk whenever $\omega_t>0$ and $\epsilon_x<\|\delta_x\|/2$.
The argmax discards the ambiguity of each posterior, since a one-hot target stays at squared distance \looseness=-1
\begin{equation}
\|e_c-\bar e_j\|^2=1-2\bar e_{jc}+\|\bar e_j\|^2\ge(1-\bar e_{jc})^2
\label{eq:onehot}
\end{equation}
from the posterior for every class $c$, which is large on ambiguous images, whereas a soft label can match the posterior. \Cref{supp:pl-dist} measures how far the soft labels of {\ours} and the hard pseudo-labels of zero-shot CLIP~\cite{radford2021clip} and N2L~\cite{li2026n2l} are from the true labels.

By \cref{lem:loo}, the LOO output of an image is the output of a classifier that has not seen it, like a test prediction, and costs no re-solving.
By (c), the criterion of \cref{eq:omega} is the expected number of correct LOO decisions plus an error term that vanishes when the soft labels are the posteriors.
Against the weight $\omega^\star$ with the most correct decisions in expectation, the selected weight then loses at most the $L_1$ distance of the soft labels to the posteriors, summed over the images that the two weights decide differently.

Parts (a) and (b) compare predictions in squared error and part (c) counts correct LOO decisions on the current images, so none of them bounds the test accuracy, and the argmax can be more accurate, as for $A_T$ on CUB in \cref{fig:ablation_posterior}.
The results hold for any soft labels computed from the observations, including those of the prior selection rule and of EM with $\tau=4$. They state what these modules should aim at, not that they reach it.
The reference prediction $F_{e_y}$ keeps the weights $m_j$ that \cref{eq:weights} computes from the soft labels.
The LOO of \cref{lem:loo} removes an image only from the ridge step, since the soft labels and weights computed with this image are kept. \looseness=-1

\subsection{Proofs}
\label{supp:theory-proofs}

\begin{proof}[Proof of \cref{thm:soft}]
Given $\mathcal{F}_t$, the labels follow a probability vector over the finitely many label configurations, $\mathbb{E}_t$ is the expectation under this vector, and every $\mathcal{F}_t$-measurable quantity is a constant under it.
We write $B_t=A_t+\lambda I_D$ and $\phi_j=\phi(x_j)$, and $j$ runs over all seen images.

\emph{Step 1: the recursion is ridge regression on all seen images.}
By induction on $t$, the recursion of \cref{eq:analytic} gives
\begin{equation}
A_t=\sum_jm_j\phi_j\phi_j^\top,\qquad C_t=\sum_jm_j\phi_jQ_j^\top.
\end{equation}
Written over all classes, $C_{t-1}$ has zero columns for the classes of task $t$, as every soft label is zero outside the classes of its own task, so the zero block of \cref{eq:analytic} loses nothing.
Hence $W_t=B_t^{-1}C_t=W(Q)$, although no seen image is stored once its task ends.

\emph{Step 2: conditional mean and risk.}
$W(\Theta)$ is linear in $\Theta$ with the $\mathcal{F}_t$-measurable coefficients $m_jB_t^{-1}\phi_j$, so $\mathbb{E}_t[W(e_y)]=W(\bar e)$, and $\mathbb{E}_t[F_{e_y}(x)]=F_{\bar e}(x)$ because $\omega_t$ and $\pzsbar(x)$ are constants.
With $o=W(\Theta)^\top\phi(x)$ and $o^\star=W(e_y)^\top\phi(x)$, the zero-shot term cancels in
\begin{equation}
F_\Theta(x)-F_{e_y}(x)=\omega_t\,(o-o^\star).
\end{equation}
As $\bar o=W(\bar e)^\top\phi(x)=\mathbb{E}_t[o^\star]$, the cross term of
\begin{equation}
\begin{aligned}
\|o-o^\star\|^2={}&\|o-\bar o\|^2+\|\bar o-o^\star\|^2\\
&+2(o-\bar o)^\top(\bar o-o^\star)
\end{aligned}
\end{equation}
has conditional mean zero, so
\begin{equation}
R_x(\Theta)=\omega_t^2\bigl(\|o-\bar o\|^2+\sigma_t^2(x)\bigr),
\label{eq:pf-bv}
\end{equation}
which is the identity of (a) and is smallest at $\Theta=\bar e$.

\emph{Step 3: target errors pass through the leverage.}
As $B_t$ is symmetric, the target errors enter the output at $x$ as
\begin{equation}
o-\bar o=\sum_jm_ja_j(\Theta_j-\bar e_j),\qquad a_j=\phi_j^\top b,
\end{equation}
where $b=B_t^{-1}\phi(x)$, so that $s_t(x)=\phi(x)^\top b$.
The triangle and Cauchy-Schwarz inequalities give
\begin{equation}
\|o-\bar o\|^2\le\Bigl(\sum_jm_ja_j^2\Bigr)\Bigl(\sum_jm_j\|\Theta_j-\bar e_j\|^2\Bigr).
\end{equation}
The inner product of $B_tb=\phi(x)$ with $b$ gives
\begin{equation}
s_t(x)=\sum_jm_ja_j^2+\lambda\|b\|^2,
\end{equation}
so $\sum_jm_ja_j^2\le s_t(x)$, which proves the upper bound of (a).
The same identity gives $\lambda\|b\|^2\le s_t(x)$, and Cauchy-Schwarz gives $s_t(x)^2\le\|\phi(x)\|^2\|b\|^2$, so
\begin{equation}
\lambda s_t(x)^2\le\|\phi(x)\|^2\,\lambda\|b\|^2\le\|\phi(x)\|^2s_t(x),
\end{equation}
and $s_t(x)\le\|\phi(x)\|^2/\lambda$ as $s_t(x)\ge0$.

\emph{Step 4: posterior versus argmax.}
The error of the soft labels at $x$ is $\epsilon_x=\|\zeta\|$ with $\zeta=(W(Q)-W(\bar e))^\top\phi(x)$, and applying \cref{eq:pf-bv} to $\hat Q$ and $Q$ gives
\begin{equation}
\begin{aligned}
R_x(\hat Q)-R_x(Q)&=\omega_t^2\bigl(\|\delta_x+\zeta\|^2-\|\zeta\|^2\bigr)\\
&=\omega_t^2\bigl(\|\delta_x\|^2+2\delta_x^\top\zeta\bigr),
\end{aligned}
\end{equation}
and $|\delta_x^\top\zeta|\le\|\delta_x\|\epsilon_x$ gives (b).
The bound on $\epsilon_x^2$ is Step~3 with $\Theta=Q$, and $Q=\bar e$ gives $\zeta=0$.
In \cref{eq:onehot}, the identity holds as $\|e_c\|=1$ and $e_c^\top\bar e_j=\bar e_{jc}$, and the inequality follows from $\|\bar e_j\|^2\ge\bar e_{jc}^2$, as $1-2\bar e_{jc}+\bar e_{jc}^2=(1-\bar e_{jc})^2$.

\emph{Step 5: the fusion criterion.}
As each decision $\hat c_i(\omega)$ is $\mathcal{F}_t$-measurable, it is correct with probability
\begin{equation}
\bar e_{i,\hat c_i(\omega)}=Q_{i,\hat c_i(\omega)}-\xi_{i,\hat c_i(\omega)},\qquad\xi_i=Q_i-\bar e_i,
\end{equation}
and summing over $i$ gives \cref{eq:thm-crit}.
This identity at $\omega_t$ and $\omega^\star$, with $J(\omega_t)\ge J(\omega^\star)$, gives
\begin{equation}
\begin{aligned}
&\mathbb{E}_t\bigl[N_{\mathrm{cor}}(\omega^\star)-N_{\mathrm{cor}}(\omega_t)\bigr]\\
&\quad\le\sum_{i\in\mathcal{S}}\bigl(\xi_{i,\hat c_i(\omega_t)}-\xi_{i,\hat c_i(\omega^\star)}\bigr),
\end{aligned}
\end{equation}
since the images decided alike contribute the same term at both weights.
Each summand involves two different classes $a$ and $b$ and is at most $|\xi_{ia}|+|\xi_{ib}|\le\|\xi_i\|_1$.
The lower bound holds because $\omega^\star$ maximizes $\mathbb{E}_t[N_{\mathrm{cor}}(\omega)]$.
\end{proof}

\begin{proof}[Proof of \cref{lem:loo}]
Setting the weight of image $i$ to zero replaces $B_t$ and $C_t$ by
\begin{equation}
B_-=B_t-m_i\phi_i\phi_i^\top,\qquad C_-=C_t-m_i\phi_iQ_i^\top,
\end{equation}
and $B_-\succeq\lambda I_D$ is positive definite.
Let $w=B_-^{-1}\phi_i$ and $h_-=m_i\phi_i^\top w\ge0$.
Then
\begin{equation}
\begin{aligned}
B_tw&=(1+h_-)\phi_i,\\
h_i&=m_i\phi_i^\top B_t^{-1}\phi_i=\frac{h_-}{1+h_-}\in[0,1),
\end{aligned}
\end{equation}
since $B_t^{-1}\phi_i=w/(1+h_-)$.
For $v=B_t^{-1}\phi_i$, we have $B_-v=(1-h_i)\phi_i$, hence $B_-^{-1}\phi_i=v/(1-h_i)$, and as $B_-$ is symmetric, the re-solved classifier $B_-^{-1}C_-$ outputs at $x_i$
\begin{equation}
\begin{aligned}
C_-^\top B_-^{-1}\phi_i&=\frac{C_t^\top v-m_i(\phi_i^\top v)\,Q_i}{1-h_i}\\
&=\frac{r_i-h_iQ_i}{1-h_i}=\tilde r_i,
\end{aligned}
\end{equation}
since $r_i=W_t^\top\phi_i=C_t^\top v$ and $h_i=m_i\phi_i^\top v$.
\end{proof}


\section{More Implementation Details}
\label{supp:impl}

This section lists the datasets and task splits of our experiments and gives the full algorithm of {\ours}.

\begin{table}[!t]
\caption{Datasets and task splits.
Train and Test are the numbers of training and test images, and B$b$Inc$k$ is a first task of $b$ classes followed by tasks of $k$ classes each.}
\label{tab:supp-data}
\centering\footnotesize
\setlength{\tabcolsep}{5.0pt}
\begin{tabular}{@{}lcccc@{}}
\toprule
Dataset & Classes & Train & Test & Settings\\
\midrule
Aircraft & 100 & 6{,}667 & 3{,}333 & B0Inc10, B50Inc10\\
Cars & 100 & 4{,}135 & 4{,}083 & B0Inc10, B50Inc10\\
CIFAR100 & 100 & 50{,}000 & 10{,}000 & B0Inc10, B50Inc10\\
CUB & 200 & 9{,}430 & 2{,}358 & B0Inc20, B100Inc20\\
ObjectNet & 200 & 26{,}509 & 6{,}628 & B0Inc20, B100Inc20\\
UCF & 100 & 10{,}553 & 2{,}639 & B0Inc10, B50Inc10\\
Food & 100 & 79{,}988 & 20{,}012 & B0Inc10, B50Inc10\\
ImageNet-R & 200 & 24{,}000 & 6{,}000 & B0Inc20, B100Inc20\\
SUN & 300 & 72{,}870 & 18{,}179 & B0Inc30, B150Inc30\\
\bottomrule
\end{tabular}
\end{table}

\begin{algorithm}[!t]
\caption{{\ours} on task $t$}
\label{alg:task}
\small
\begin{algorithmic}[1]
\Require class names of $\mathcal{Y}_{1:t}$, features $X_t\in\mathbb{R}^{n_t\times d}$, statistics $(G,g,N,H,\nu)$ and $(A,C)$, fixed $R$
\Ensure classifier $W_t$, fusion weight $\omega_t$, updated statistics
\State Text embeddings $z_c$ from the first prompt template, $c\in\mathcal{Y}_{1:t}$
\State $\Praw$ by \cref{eq:zs}, $\Pbal$ by Sinkhorn (\cref{eq:sinkhorn})
\State $q^{\mathrm{bal}}\gets\textsc{EM}(X_t,\Pbal,1)$
\State $\rho_t$ from the marginals of $\Praw$ and $q^{\mathrm{bal}}$, select $\Psel_t$ (\cref{eq:rho})
\State $q\gets\textsc{EM}(X_t,\Psel_t,\tau)$ with $\tau=4$
\State Update $(G,g,N,H,\nu)$ with $X_t$ and $q$
\State $\Phi\gets\max(X_tR,0)$, weights $m$ by \cref{eq:weights}
\State Update $(A,C)$ and solve $W_t$ by Cholesky, $\lambda=10$ (\cref{eq:analytic})
\State LOO predictions (\cref{eq:loo}), $\omega_t$ by \cref{eq:omega} with $\pzs$ from $z_c$, $c\in\mathcal{Y}_{1:t}$
\State Discard $X_t$, $\Phi$ and $q$
\Statex \textbf{Inference:} prompt-ensemble embeddings $\bar z_c$, $c\in\mathcal{Y}_{1:t}$, predict with \cref{eq:fusion}
\Function{EM}{$X,P,\tau$}
\State $\bar x$ and $V$ ($d_p=128$) from the statistics merged with $X$
\State $q\gets P$
\For{pass $=1,2$}
\State $\Sw$ by \cref{eq:sw}, OAS shrinkage, Cholesky factor $L$, whitened $\tilde x$
\State Alternate the M-step and the E-step of \cref{eq:em} for at most 21 iterations, until the mean of $\log Z_i$, with $Z_i$ the denominator of the E-step, changes by less than $10^{-5}$
\EndFor
\State \Return $q$
\EndFunction
\end{algorithmic}
\end{algorithm}

\begin{table*}[!t]
\caption{Comparison on Food, ImageNet-R and SUN.
$\bar A$ and $A_T$ in \%, and Mean averages the six settings.
}
\label{tab:supp-more}
\centering\footnotesize
\setlength{\tabcolsep}{5pt}
\begin{tabular}{@{}llcccccccccccccc@{}}
\toprule
\multirow{3}{*}[-1.25ex]{Method} & \multirow{3}{*}[-1.25ex]{Venue} & \multicolumn{4}{c}{Food} & \multicolumn{4}{c}{ImageNet-R} & \multicolumn{4}{c}{SUN} & \multicolumn{2}{c}{\multirow{2}{*}[-0.7ex]{Mean}}\\
\cmidrule(lr){3-6}\cmidrule(lr){7-10}\cmidrule(lr){11-14}
 & & \multicolumn{2}{c}{B0Inc10} & \multicolumn{2}{c}{B50Inc10} & \multicolumn{2}{c}{B0Inc20} & \multicolumn{2}{c}{B100Inc20} & \multicolumn{2}{c}{B0Inc30} & \multicolumn{2}{c}{B150Inc30} & \multicolumn{2}{c}{}\\
\cmidrule(lr){3-4}\cmidrule(lr){5-6}\cmidrule(lr){7-8}\cmidrule(lr){9-10}\cmidrule(lr){11-12}\cmidrule(lr){13-14}\cmidrule(l){15-16}
 & & $\bar A$ & $A_T$ & $\bar A$ & $A_T$ & $\bar A$ & $A_T$ & $\bar A$ & $A_T$ & $\bar A$ & $A_T$ & $\bar A$ & $A_T$ & $\bar A$ & $A_T$\\
\midrule
ZS-CLIP & ICML'21 & 87.86 & 81.99 & 84.78 & 81.99 & 83.12 & 76.62 & 79.23 & 76.62 & 79.45 & 72.14 & 74.99 & 72.14 & 81.57 & 76.92\\
ENGINE & ICCV'25 & 88.30 & 81.78 & 84.72 & 81.55 & 84.14 & 76.93 & 79.26 & 76.56 & 83.04 & 76.17 & 78.06 & 75.10 & 82.92 & 78.02\\
N2L & ICLR'26 & \underline{90.58} & \underline{85.50} & \underline{87.18} & \underline{85.01} & \underline{86.00} & \underline{81.06} & \underline{81.89} & \underline{80.29} & \underline{84.38} & \underline{78.85} & \underline{79.15} & \underline{77.40} & \underline{84.86} & \underline{81.35}\\
{\ours} & Ours & \textbf{91.13} & \textbf{86.25} & \textbf{87.94} & \textbf{85.66} & \textbf{86.86} & \textbf{81.48} & \textbf{82.82} & \textbf{80.99} & \textbf{85.46} & \textbf{80.14} & \textbf{80.58} & \textbf{78.85} & \textbf{85.80} & \textbf{82.23}\\
\midrule
\rowcolor{labelrow}\labelrowfirst{ENGINE+Label} & ICCV'25 & 88.52 & 81.85 & 85.08 & 81.59 & 84.61 & 77.12 & 79.96 & 76.78 & 83.51 & 76.21 & 78.83 & 75.20 & 83.42 & \labelrowlast{78.13}\\
\rowcolor{labelrow}\labelrowfirst{N2L+Label} & ICLR'26 & 90.95 & 85.84 & 88.38 & 85.84 & 86.61 & 81.52 & 83.88 & 81.51 & 87.42 & 81.50 & 84.31 & 81.50 & 86.93 & \labelrowlast{82.95}\\
\rowcolor{labelrow}\labelrowfirst{{\ours}+Label} & Ours & 91.36 & 86.49 & 88.84 & 86.50 & 87.49 & 82.07 & 84.32 & 82.12 & 86.72 & 80.63 & 83.48 & 80.64 & 87.04 & \labelrowlast{83.08}\\
\bottomrule
\end{tabular}
\end{table*}

\begin{table*}[!t]
\caption{Comparison with exemplar-based methods on six datasets.
$\bar A$ and $A_T$ in \%, and Mean averages the six settings of each block.
CLAP4CLIP and PROOF are trained on zero-shot pseudo-labels and store 20 images per class of earlier tasks, while ZS-CLIP and {\ours} store none.
}
\label{tab:supp-exemplar}
\centering\footnotesize
\setlength{\tabcolsep}{5pt}
\begin{tabular}{@{}llcccccccccccccc@{}}
\toprule
\multirow{3}{*}[-1.25ex]{Method} & \multirow{3}{*}[-1.25ex]{Venue} & \multicolumn{4}{c}{Aircraft} & \multicolumn{4}{c}{Cars} & \multicolumn{4}{c}{CIFAR100} & \multicolumn{2}{c}{\multirow{2}{*}[-0.7ex]{Mean}}\\
\cmidrule(lr){3-6}\cmidrule(lr){7-10}\cmidrule(lr){11-14}
 & & \multicolumn{2}{c}{B0Inc10} & \multicolumn{2}{c}{B50Inc10} & \multicolumn{2}{c}{B0Inc10} & \multicolumn{2}{c}{B50Inc10} & \multicolumn{2}{c}{B0Inc10} & \multicolumn{2}{c}{B50Inc10} & \multicolumn{2}{c}{}\\
\cmidrule(lr){3-4}\cmidrule(lr){5-6}\cmidrule(lr){7-8}\cmidrule(lr){9-10}\cmidrule(lr){11-12}\cmidrule(lr){13-14}\cmidrule(l){15-16}
 & & $\bar A$ & $A_T$ & $\bar A$ & $A_T$ & $\bar A$ & $A_T$ & $\bar A$ & $A_T$ & $\bar A$ & $A_T$ & $\bar A$ & $A_T$ & $\bar A$ & $A_T$\\
\midrule
ZS-CLIP & ICML'21 & 26.61 & 17.16 & 21.66 & 17.16 & 82.90 & 76.73 & 78.74 & 76.73 & 81.81 & 71.38 & 76.49 & 71.38 & 61.37 & 55.09\\
CLAP4CLIP & NeurIPS'24 & \underline{41.15} & \underline{35.81} & 26.88 & \underline{28.43} & \underline{90.02} & \underline{86.46} & 79.68 & 75.88 & \underline{86.34} & \underline{79.11} & \underline{82.90} & \underline{80.57} & \underline{67.83} & \underline{64.38}\\
PROOF & TPAMI'25 & 38.10 & 33.62 & \underline{28.10} & 28.15 & 89.23 & 85.09 & \underline{83.83} & \underline{83.03} & 84.85 & 77.47 & 79.17 & 76.36 & 67.21 & 63.95\\
{\ours} & Ours & \textbf{51.35} & \textbf{49.70} & \textbf{36.75} & \textbf{39.80} & \textbf{92.73} & \textbf{89.40} & \textbf{88.53} & \textbf{87.95} & \textbf{88.82} & \textbf{82.71} & \textbf{83.87} & \textbf{81.49} & \textbf{73.68} & \textbf{71.84}\\
\midrule[\heavyrulewidth]
\multirow{3}{*}[-1.25ex]{Method} & \multirow{3}{*}[-1.25ex]{Venue} & \multicolumn{4}{c}{CUB} & \multicolumn{4}{c}{ObjectNet} & \multicolumn{4}{c}{UCF} & \multicolumn{2}{c}{\multirow{2}{*}[-0.7ex]{Mean}}\\
\cmidrule(lr){3-6}\cmidrule(lr){7-10}\cmidrule(lr){11-14}
 & & \multicolumn{2}{c}{B0Inc20} & \multicolumn{2}{c}{B100Inc20} & \multicolumn{2}{c}{B0Inc20} & \multicolumn{2}{c}{B100Inc20} & \multicolumn{2}{c}{B0Inc10} & \multicolumn{2}{c}{B50Inc10} & \multicolumn{2}{c}{}\\
\cmidrule(lr){3-4}\cmidrule(lr){5-6}\cmidrule(lr){7-8}\cmidrule(lr){9-10}\cmidrule(lr){11-12}\cmidrule(lr){13-14}\cmidrule(l){15-16}
 & & $\bar A$ & $A_T$ & $\bar A$ & $A_T$ & $\bar A$ & $A_T$ & $\bar A$ & $A_T$ & $\bar A$ & $A_T$ & $\bar A$ & $A_T$ & $\bar A$ & $A_T$\\
\midrule
ZS-CLIP & ICML'21 & 75.47 & 63.72 & 69.06 & 63.72 & 38.43 & 26.43 & 31.12 & 26.43 & 75.88 & 67.79 & 71.68 & 67.79 & 60.27 & 52.65\\
CLAP4CLIP & NeurIPS'24 & \underline{82.45} & \underline{74.85} & \underline{75.74} & \underline{72.71} & \underline{49.00} & \underline{38.55} & \underline{41.85} & \underline{37.77} & 87.10 & 83.82 & \underline{81.38} & \underline{83.50} & \underline{69.59} & \underline{65.20}\\
PROOF & TPAMI'25 & 80.50 & 73.98 & 73.74 & 71.83 & 45.46 & 34.93 & 35.36 & 32.16 & \underline{89.41} & \underline{86.00} & 81.23 & 81.76 & 67.62 & 63.44\\
{\ours} & Ours & \textbf{86.71} & \textbf{80.38} & \textbf{78.76} & \textbf{77.21} & \textbf{53.06} & \textbf{41.62} & \textbf{41.88} & \textbf{37.78} & \textbf{95.92} & \textbf{94.72} & \textbf{89.35} & \textbf{90.40} & \textbf{74.28} & \textbf{70.35}\\
\bottomrule
\end{tabular}
\end{table*}

\paragraph{Datasets and task splits.}
\label{supp:data}
\Cref{tab:supp-data} lists the class subsets, the image counts and the two settings of each dataset. The counts are those of the features extracted in our runs, which use the train and test splits of N2L~\cite{li2026n2l}.

\paragraph{Algorithm.}
\label{supp:alg}
\Cref{alg:task} gives the steps that {\ours} runs on task $t$.
Lines 1 and 2 compute the raw zero-shot distribution $\Praw$ with the first prompt template and its debiased version $\Pbal$ with Sinkhorn iterations.
Lines 3 and 4 form the prior selection rule, which runs EM from $\Pbal$ without tempering and keeps $\Praw$ only if the images reproduce its class imbalance.
Line 5 runs EM again from the selected prior with $\tau=4$, and line 6 adds the images and their soft labels to the whitening statistics.
Lines 7 to 9 fit the analytic classifier to the soft labels with class-balancing weights and select $\omega_t$ from its closed-form LOO outputs.
Line 10 then discards the features and soft labels of the task, so later tasks see it only through the accumulated statistics.
Each EM call computes the mean and the principal directions once and makes two passes, each re-estimating the within-class covariance and the whitening before the vMF iterations.
Because EM only reads the statistics, the EM run of the prior selection rule leaves no trace in them.

\section{Additional Experiments}
\label{supp:results}

\begin{figure*}[!t]
\setlength{\abovecaptionskip}{0.2cm}
    \centering
	\begin{overpic}[width=\textwidth]{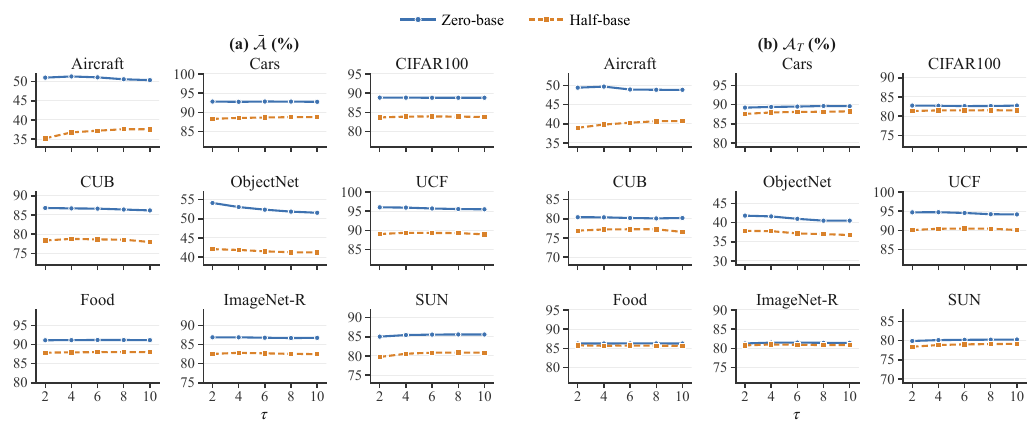}
    \end{overpic}
    \caption{Sensitivity to the temperature $\tau$ of the final EM.
(a)~$\bar A$ and (b)~$A_T$ (\%) on the zero-base and half-base settings of the nine datasets for $\tau\in\{2,4,6,8,10\}$, with all other hyperparameters fixed.
{\ours} uses $\tau=4$, and each point is the mean over three runs.}
 \label{fig:tau_sensitivity}
\end{figure*}

\subsection{More Comparisons}
\label{supp:comparisons}

\paragraph{Food, ImageNet-R and SUN.}
\label{supp:more}
\Cref{tab:supp-more} completes \cref{tab:main} with the three datasets that N2L~\cite{li2026n2l} reports in its Tab.~6, which gives only ZS-CLIP~\cite{radford2021clip}, ENGINE~\cite{zhou2025engine}, N2L, ENGINE+Label and N2L+Label for them.
{\ours} is the best of the methods trained without labels in both metrics on all six settings, and its gains over N2L range from 0.42\% to 1.45\% on these datasets, where the refined pseudo-labels of N2L are between 85.70\% and 96.22\% correct.
{\ours}+Label is above {\ours} by 0.23\% to 2.90\% in $\bar A$ and below N2L+Label in both metrics on both settings of SUN, the dataset with the most classes.
In the Mean columns, {\ours} is 0.94\% and 0.88\% above N2L in $\bar A$ and $A_T$, far less than on the two blocks of \cref{tab:main}, and ground-truth labels add only 1.24\% and 0.85\% to its Mean.

\paragraph{Exemplar-based methods.}
\label{supp:exemplar}
\Cref{tab:supp-exemplar} compares {\ours} with two exemplar-based methods, PROOF~\cite{zhou2025proof} and CLAP4CLIP~\cite{jha2024clap4clip}, which keep 20 training images per class of earlier tasks.
N2L ran both in CIL with unlabeled data, training them on pseudo-labels from zero-shot CLIP, as its other baselines, and their results, as well as those of ZS-CLIP, are quoted from Table~4 of~\cite{li2026n2l}, which covers the 12 settings of Aircraft, Cars, CIFAR100, CUB, ObjectNet and UCF.
{\ours} stores no image and is above both methods in $\bar A$ and $A_T$ on all 12 settings, although only by 0.03\% and 0.01\% on ObjectNet B100Inc20, where it is level with CLAP4CLIP.
Averaged over the 12 settings, it reaches 73.98\% in $\bar A$ and 71.10\% in $A_T$, against 68.71\% and 64.79\% for CLAP4CLIP, the stronger of the two exemplar-based methods.
The Mean columns show this margin on both blocks, 5.85\% and 4.69\% in $\bar A$ and 7.46\% and 5.15\% in $A_T$ over CLAP4CLIP.

\subsection{Robustness to Hyperparameters and Backbones}
\label{supp:robust}

\paragraph{Temperature $\tau$.}
\label{sec:exp-tau}
\Cref{fig:tau_sensitivity} reports $\bar A$ and $A_T$ on the 18 settings when the temperature $\tau$ of the final EM of every task takes the values 2, 4, 6, 8 and 10.
All other hyperparameters are those of \cref{sec:exp-setup}, the debiased run of the prior selection rule keeps $\tau=1$ (\cref{sec:prior}), and each point is the mean of three runs with seeds 1993, 1994 and 1995 at that value of $\tau$.
Over this range, $\bar A$ and $A_T$ each vary by less than 1\% on 15 of the 18 settings, and $\bar A$ varies by at most 0.18\% on CIFAR100 B0Inc10, Cars B0Inc10, ImageNet-R B0Inc20 and both Food settings.
Averaged over the 18 settings, $\tau=4$ gives the highest $\bar A$ and $A_T$ of the five values, 77.92\% and 74.81\%, and the other four values are lower by at most 0.25\% and 0.18\%.
On single settings, $\tau=4$ is within 0.30\% of the best of the five values on 16 of the 18 settings in each metric.
Choosing $\tau$ for each dataset by the mean $\bar A$ of the other eight datasets selects $\tau=4$ for every dataset except ObjectNet, where it selects $\tau=6$, and lowers the mean $\bar A$ over the 18 settings by 0.06\%.
A single $\tau$ shared by all datasets therefore costs little compared with a per-dataset choice, and it is chosen without labeled data.

\begin{figure*}[!t]
\setlength{\abovecaptionskip}{0.2cm}
    \centering
	\begin{overpic}[width=\textwidth]{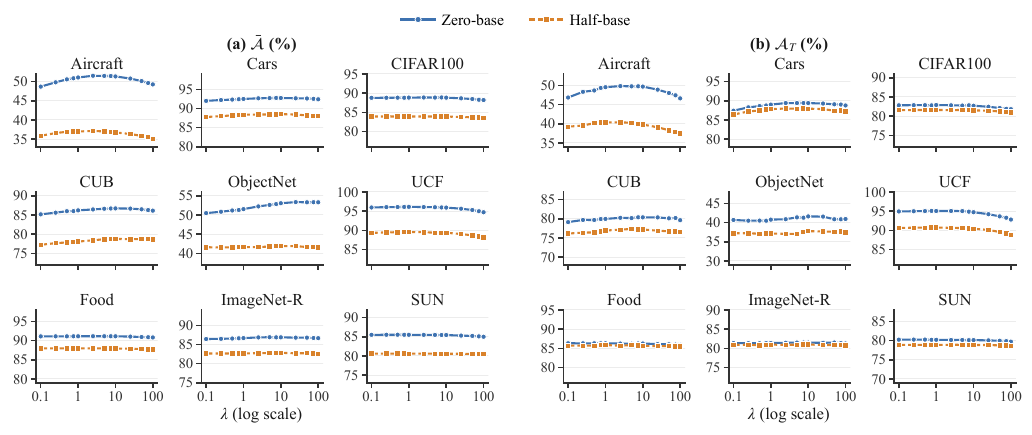}
    \end{overpic}
    \caption{Sensitivity to the ridge coefficient $\lambda$.
(a)~$\bar A$ and (b)~$A_T$ (\%) on the zero-base and half-base settings of the nine datasets for 13 values of $\lambda$ from 0.1 to 100, with all other hyperparameters fixed.
{\ours} uses $\lambda=10$, and each point is the mean over three runs.}
 \label{fig:lambda_sensitivity}
\end{figure*}

For fixed class directions and concentration, dividing the exponent of \cref{eq:em} by $\tau$ divides by $\tau$ every shift that the images apply to the log-odds of the prior $\Psel_t$.
With a small $\tau$, small differences between cosines, scaled by the large fitted concentration, override the prior.
The soft labels then become sharp and follow the vMF clusters even where these clusters do not match the classes, and they enter the statistics of the closed-form classifier without later revision.
With a large $\tau$, every $q_i$ moves toward $\Psel_t$, and the soft labels become flatter and keep fewer of the corrections that the images make to the zero-shot prior.
The mean $\bar A$ of the nine zero-base settings, whose tasks hold 10 to 30 classes, decreases from 81.40\% at $\tau=2$ to 80.95\% at $\tau=10$.
The mean $\bar A$ of the nine half-base settings, whose first task holds 50 to 150 classes, is 74.08\% at $\tau=2$ and between 74.39\% and 74.53\% for $\tau\ge4$.
This is consistent with more pairs of similar classes in an EM run over more classes, where strong image evidence moves more images to a wrong class.
On ObjectNet B0Inc20, $\bar A$ falls from 54.13\% at $\tau=2$ to 51.57\% at $\tau=10$, and $\tau=4$ is 1.07\% below the best value.
On Aircraft B50Inc10, whose first task holds 50 aircraft variants, $\bar A$ rises from 35.22\% at $\tau=2$ to 37.59\% at $\tau=8$, 0.84\% above $\tau=4$, and $A_T$ is 0.92\% higher at $\tau=10$ than at $\tau=4$.
SUN B150Inc30 also prefers $\tau\ge6$, by at most 0.30\% in $\bar A$ and 0.32\% in $A_T$.
These settings would gain from a $\tau$ chosen per dataset, but such a choice needs labeled data from each dataset.

\paragraph{Ridge coefficient $\lambda$.}
\label{supp:lambda}
\Cref{fig:lambda_sensitivity} reports $\bar A$ and $A_T$ on the 18 settings for 13 values of the ridge coefficient $\lambda$ of \cref{eq:analytic} from 0.1 to 100, with $\tau=4$ and all other hyperparameters as in \cref{sec:exp-setup}, and each point is the mean of three runs with seeds 1993, 1994 and 1995.
The coefficient $\lambda$ enters only the solve of $W_t$ and the leverages of \cref{eq:loo}. The soft labels and the weights $m_i$ are the same for every $\lambda$, and only $W_t$ and $\omega_t$ change.
Averaged over the 18 settings, $\bar A$ rises from 77.28\% at $\lambda=0.1$, the value used by N2L~\cite{li2026n2l}, to between 77.86\% and 77.93\% for $\lambda$ from 2.5 to 10, and falls to 77.34\% at $\lambda=100$.
$A_T$ follows the same course, from 74.25\% to between 74.80\% and 74.83\% and down to 73.95\%.
For $\lambda$ from 1 to 10, $\bar A$ and $A_T$ each vary by less than 0.5\% on 15 of the 18 settings.

\begin{figure*}[!t]
\setlength{\abovecaptionskip}{0.2cm}
    \centering
	\begin{overpic}[width=\textwidth]{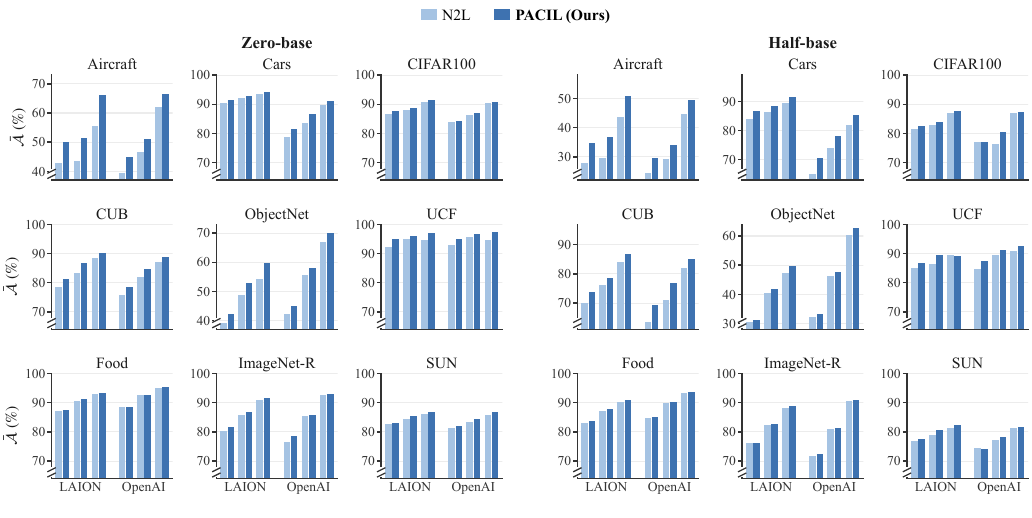}
	    \put(0,488){\small (a)}
    \end{overpic}\\[1mm]
	\begin{overpic}[width=\textwidth]{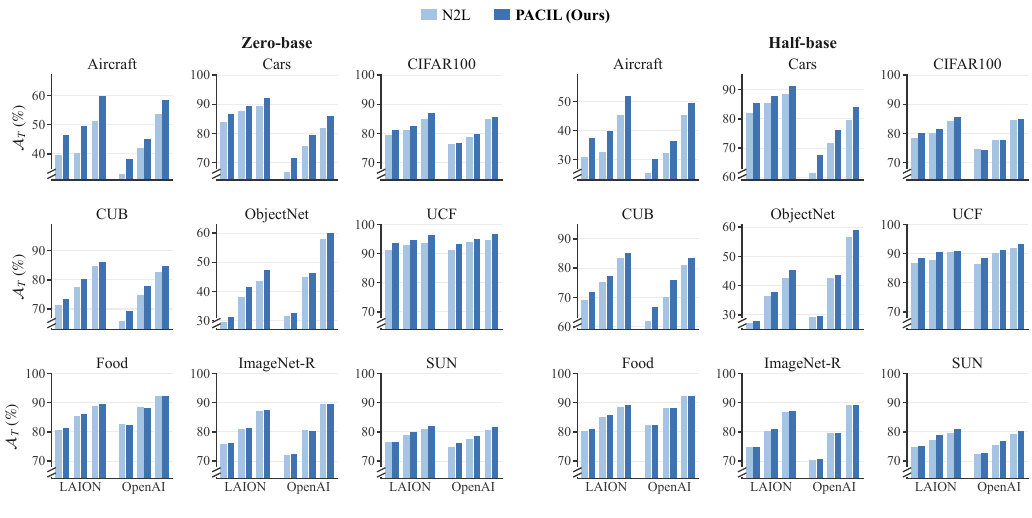}
	    \put(0,488){\small (b)}
    \end{overpic}
    \caption{CLIP backbones.
(a)~$\bar A$ and (b)~$A_T$ (\%) of N2L and {\ours} with ViT-B/32, ViT-B/16 and ViT-L/14 (left to right within each group), each with LAION-400M and OpenAI weights, on the nine datasets in the zero-base (left) and half-base (right) settings.
Both methods are re-run in the same setting.}
 \label{fig:backbones}
\end{figure*}

$\lambda$ is the weight of the penalty $\lambda\|W\|_F^2$ of a weighted ridge regression whose data terms have weights $m_i$ summing to about $n_t$ per task.
A small $\lambda$ barely damps the directions of $A_t$ with small eigenvalues, so $W_t$ follows the soft label of each image closely.
It also raises the leverages $h_i$, and the LOO predictions that select $\omega_t$ then amplify the fitted residuals by a larger factor $1/(1-h_i)$.
From $\lambda=10$ to 0.1, $\bar A$ falls by 2.66\% on Aircraft B0Inc10 and 2.58\% on ObjectNet B0Inc20.
A large $\lambda$ shrinks $W_t$ most along the directions of $A_t$ with small eigenvalues, and $\lambda W_t$ tends to $C_t$, a sum of features weighted by the soft labels that ignores the correlations in $A_t$, which can remove the fine distinctions between similar classes.
From $\lambda=10$ to 100, $\bar A$ falls on 17 of the 18 settings and $A_T$ on all 18, most on Aircraft, by 2.08\% and 1.62\% in $\bar A$, and on UCF, by 1.21\% and 1.17\%, whose $\bar A$ peaks at $\lambda=2.5$ and $\lambda=1$.
ObjectNet B0Inc20 is the only exception, as it rises to 53.35\% at $\lambda=25$ and stays within 0.05\% of this value up to $\lambda=100$. \looseness=-1

We use $\lambda=10$ because it lies on the plateau of both means and loses least on the settings that depart most from it.
Its mean $\bar A$ and $A_T$ are 0.01\% and 0.02\% below the best of the 13 values, reached at $\lambda=7.5$ and at $\lambda=5$ and 7.5.
Among the 13 values it has the smallest largest shortfall in $\bar A$ from the best value of a setting, 0.38\% on Aircraft B50Inc10, and it is within 0.20\% of that best value on 16 settings in $\bar A$ and on 15 in $A_T$, where its largest shortfall is 0.58\%, also on Aircraft B50Inc10.
As the weights $m_i$ sum to about $n_t$ per task in both methods, this value also compares directly with the 0.1 of N2L (\cref{sec:head}).

\begin{table*}[!t]
\caption{Comparison under noisy task boundaries. $\bar A$ and $A_T$ in \%, and Mean averages the six settings of each block.
From the second task on, a fraction $p$ of the unlabeled training images of each task comes from earlier tasks, and the results without noise are in \cref{tab:main}.
}
\label{tab:supp-noisy}
\centering\footnotesize
\setlength{\tabcolsep}{5pt}
\begin{tabular}{@{}llcccccccccccccc@{}}
\toprule
\multirow{3}{*}[-1.25ex]{$p$} & \multirow{3}{*}[-1.25ex]{Method} & \multicolumn{4}{c}{Aircraft} & \multicolumn{4}{c}{Cars} & \multicolumn{4}{c}{CIFAR100} & \multicolumn{2}{c}{\multirow{2}{*}[-0.7ex]{Mean}}\\
\cmidrule(lr){3-6}\cmidrule(lr){7-10}\cmidrule(lr){11-14}
 & & \multicolumn{2}{c}{B0Inc10} & \multicolumn{2}{c}{B50Inc10} & \multicolumn{2}{c}{B0Inc10} & \multicolumn{2}{c}{B50Inc10} & \multicolumn{2}{c}{B0Inc10} & \multicolumn{2}{c}{B50Inc10} & \multicolumn{2}{c}{}\\
\cmidrule(lr){3-4}\cmidrule(lr){5-6}\cmidrule(lr){7-8}\cmidrule(lr){9-10}\cmidrule(lr){11-12}\cmidrule(lr){13-14}\cmidrule(l){15-16}
 & & $\bar A$ & $A_T$ & $\bar A$ & $A_T$ & $\bar A$ & $A_T$ & $\bar A$ & $A_T$ & $\bar A$ & $A_T$ & $\bar A$ & $A_T$ & $\bar A$ & $A_T$\\
\midrule
\multirow{4}{*}{20\%} & RAIL & 35.19 & 31.71 & 23.06 & 24.60 & 87.95 & 82.81 & 82.37 & 81.02 & 87.01 & 79.95 & 81.10 & 78.38 & 66.11 & 63.08\\
 & ENGINE & 33.95 & 24.72 & 26.51 & 22.56 & 86.45 & 78.32 & 82.16 & 79.18 & 84.72 & 76.61 & 79.52 & 75.86 & 65.55 & 59.54\\
 & N2L & \underline{43.11} & \underline{38.70} & \underline{29.13} & \underline{30.51} & \underline{91.06} & \underline{85.26} & \underline{86.00} & \underline{84.23} & \underline{87.68} & \underline{80.81} & \underline{82.78} & \underline{80.07} & \underline{69.96} & \underline{66.60}\\
 & {\ours} & \textbf{49.50} & \textbf{46.96} & \textbf{36.35} & \textbf{39.19} & \textbf{91.29} & \textbf{88.37} & \textbf{88.24} & \textbf{87.76} & \textbf{88.39} & \textbf{82.22} & \textbf{83.73} & \textbf{81.35} & \textbf{72.92} & \textbf{70.98}\\
\midrule
\multirow{4}{*}{50\%} & RAIL & 33.46 & 28.83 & 21.97 & 21.81 & 85.56 & 78.74 & 81.52 & 78.99 & 86.55 & 79.20 & 80.67 & 77.69 & 64.96 & 60.88\\
 & ENGINE & 33.36 & 23.25 & 25.63 & 21.39 & 85.44 & 76.56 & 80.83 & 75.90 & 83.72 & 75.68 & 78.64 & 74.50 & 64.60 & 57.88\\
 & N2L & \underline{40.49} & \underline{36.00} & \underline{27.45} & \underline{27.75} & \underline{89.32} & \underline{81.61} & \underline{85.10} & \underline{81.58} & \underline{87.02} & \underline{80.27} & \underline{82.39} & \underline{79.23} & \underline{68.63} & \underline{64.41}\\
 & {\ours} & \textbf{43.16} & \textbf{40.58} & \textbf{32.90} & \textbf{34.14} & \textbf{89.55} & \textbf{85.73} & \textbf{86.74} & \textbf{84.63} & \textbf{87.13} & \textbf{80.95} & \textbf{82.89} & \textbf{79.88} & \textbf{70.40} & \textbf{67.65}\\
\midrule[\heavyrulewidth]
\multirow{3}{*}[-1.25ex]{$p$} & \multirow{3}{*}[-1.25ex]{Method} & \multicolumn{4}{c}{CUB} & \multicolumn{4}{c}{ObjectNet} & \multicolumn{4}{c}{UCF} & \multicolumn{2}{c}{\multirow{2}{*}[-0.7ex]{Mean}}\\
\cmidrule(lr){3-6}\cmidrule(lr){7-10}\cmidrule(lr){11-14}
 & & \multicolumn{2}{c}{B0Inc20} & \multicolumn{2}{c}{B100Inc20} & \multicolumn{2}{c}{B0Inc20} & \multicolumn{2}{c}{B100Inc20} & \multicolumn{2}{c}{B0Inc10} & \multicolumn{2}{c}{B50Inc10} & \multicolumn{2}{c}{}\\
\cmidrule(lr){3-4}\cmidrule(lr){5-6}\cmidrule(lr){7-8}\cmidrule(lr){9-10}\cmidrule(lr){11-12}\cmidrule(lr){13-14}\cmidrule(l){15-16}
 & & $\bar A$ & $A_T$ & $\bar A$ & $A_T$ & $\bar A$ & $A_T$ & $\bar A$ & $A_T$ & $\bar A$ & $A_T$ & $\bar A$ & $A_T$ & $\bar A$ & $A_T$\\
\midrule
\multirow{4}{*}{20\%} & RAIL & 80.25 & 71.61 & 72.72 & 70.02 & 38.61 & 33.66 & 30.29 & 28.74 & 88.54 & 88.33 & 80.45 & 83.21 & 65.14 & 62.60\\
 & ENGINE & 75.68 & 63.60 & 70.20 & 64.19 & 44.01 & 30.34 & 34.28 & 29.39 & 86.94 & 84.54 & 82.04 & 80.64 & 65.53 & 58.78\\
 & N2L & \underline{82.10} & \underline{74.28} & \underline{74.46} & \underline{71.76} & \underline{48.10} & \underline{40.84} & \underline{40.96} & \underline{37.73} & \underline{93.15} & \underline{91.21} & \underline{85.85} & \underline{86.43} & \underline{70.77} & \underline{67.04}\\
 & {\ours} & \textbf{84.26} & \textbf{78.16} & \textbf{78.51} & \textbf{76.36} & \textbf{53.28} & \textbf{41.57} & \textbf{42.88} & \textbf{39.33} & \textbf{95.53} & \textbf{93.85} & \textbf{89.06} & \textbf{89.90} & \textbf{73.92} & \textbf{69.86}\\
\midrule
\multirow{4}{*}{50\%} & RAIL & 78.38 & 68.45 & 71.20 & 66.33 & 34.48 & 29.53 & 28.09 & 24.98 & 86.31 & 85.75 & 78.44 & 79.58 & 62.82 & 59.10\\
 & ENGINE & 74.39 & 61.79 & 69.22 & 62.82 & 42.72 & 29.75 & 33.53 & 28.41 & 84.28 & 82.83 & 80.75 & 78.63 & 64.15 & 57.37\\
 & N2L & \underline{79.86} & \underline{70.52} & \underline{72.80} & \underline{68.54} & \underline{46.48} & \underline{37.09} & \underline{39.05} & \underline{34.94} & \underline{90.80} & \underline{88.86} & \underline{84.75} & \underline{84.99} & \underline{68.96} & \underline{64.16}\\
 & {\ours} & \textbf{80.08} & \textbf{72.97} & \textbf{76.04} & \textbf{73.28} & \textbf{47.64} & \textbf{37.16} & \textbf{41.17} & \textbf{36.90} & \textbf{91.26} & \textbf{89.36} & \textbf{87.89} & \textbf{88.66} & \textbf{70.68} & \textbf{66.39}\\
\bottomrule
\end{tabular}
\end{table*}

\paragraph{CLIP backbones.}
\label{supp:backbones}
\Cref{fig:backbones} compares N2L and {\ours} on all 18 settings with ViT-B/32, ViT-B/16 and ViT-L/14, each with LAION-400M weights (OpenCLIP \texttt{laion400m\_e32}) and with OpenAI weights.
Both methods are re-run with the same code, splits and seeds, and only the backbone changes. {\ours} keeps $\tau=4$, $\lambda=10$, $d_p=128$ and $D=5000$ on every backbone without tuning, and N2L keeps its own hyperparameters.
{\ours} is above N2L in $\bar A$ on 105 of the 108 pairs of backbone and setting, and in $A_T$ on 98.
Averaged over the 18 settings, its gain in $\bar A$ is 2.10\%, 2.24\% and 2.19\% on the LAION backbones from B/32 to L/14 and 2.25\%, 2.24\% and 1.74\% on the OpenAI backbones, and its gain in $A_T$ lies between 1.78\% and 2.39\% on all six.
The gain therefore does not shrink with scale, except on OpenAI ViT-L/14, where N2L is most accurate, with a mean $\bar A$ of 82.07\%.

\begin{figure*}[!t]
\setlength{\abovecaptionskip}{0.2cm}
    \centering
	\begin{overpic}[width=\textwidth]{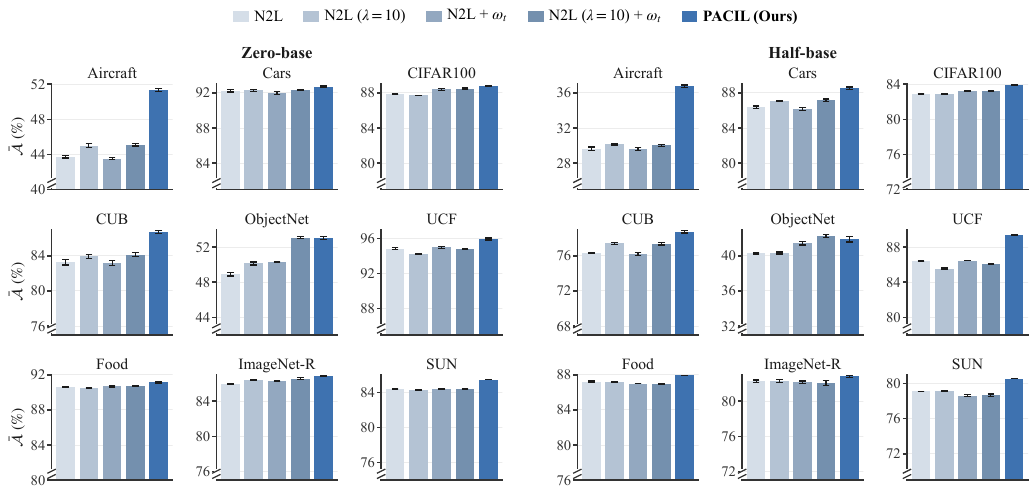}
	    \put(0,466){\small (a)}
    \end{overpic}\\[1mm]
	\begin{overpic}[width=\textwidth]{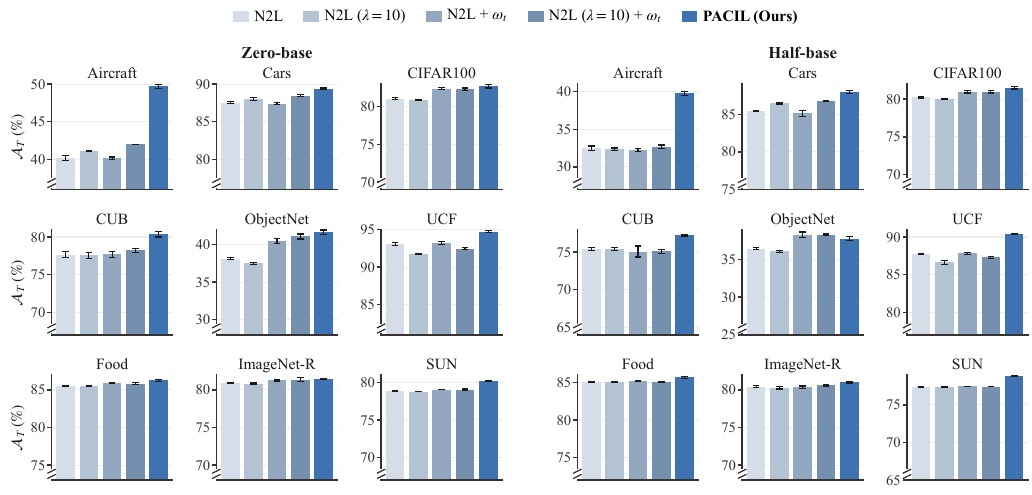}
	    \put(0,466){\small (b)}
    \end{overpic}
    \caption{N2L with components of {\ours}.
(a)~$\bar A$ and (b)~$A_T$ (\%) of N2L, N2L with the ridge coefficient $\lambda=10$, N2L with the LOO choice of the fusion weight of \cref{eq:omega}, N2L with both, and {\ours}, on the nine datasets in the zero-base (left) and half-base (right) settings.
N2L is our reproduction with its official code~\cite{li2026n2l}, with $\lambda=0.1$ and a fixed $\omega=0.8$, and its three variants are run with the same code.
Bars are means over three seeds, error bars show standard deviation, and the axes do not start at zero.}
 \label{fig:n2l_variants}
\end{figure*}

\begin{figure*}[!t]
\setlength{\abovecaptionskip}{0.2cm}
    \centering
	\begin{overpic}[width=\textwidth]{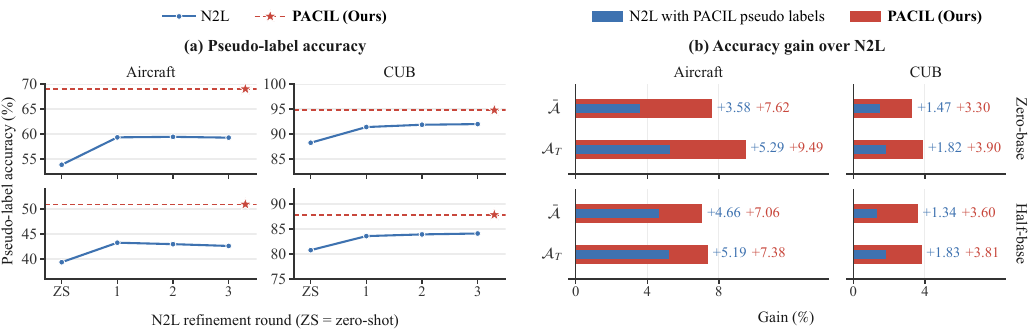}
    \end{overpic}
    \caption{Pseudo-labels of {\ours} and N2L~\cite{li2026n2l} on Aircraft and CUB.
(a)~Pseudo-label accuracy (\%) on the training images, pooled over all tasks, for N2L at zero-shot (ZS) and after each refinement round and for the argmax of the final soft labels of {\ours}, with zero-base settings on top and half-base settings at the bottom.
(b)~Gain over N2L in $\bar A$ and $A_T$ (\%) of N2L with its pseudo-labels replaced by the argmax of the {\ours} soft labels, and of {\ours}.}
 \label{fig:pseudo_label_pacil_vs_n2l}
\end{figure*}

A larger backbone gives better zero-shot predictions, which raises the accuracy of N2L and leaves less to correct where accuracy is high.
With OpenAI weights, the gain in $\bar A$ falls from B/32 to L/14 from 0.23\% to 0.07\% on Food B0Inc10 and from 2.10\% to 0.32\% on ImageNet-R B0Inc20.
On datasets that stay hard at every scale it holds or grows, as on ObjectNet B100Inc20, from 0.44\% to 2.39\% with LAION weights and from 1.04\% to 2.40\% with OpenAI weights, which is consistent with EM gaining from better image features while the zero-shot labels leave errors to correct.
Aircraft B0Inc10 shows both cases, as the gain rises from 6.91\% to 10.34\% with LAION weights and falls from 5.41\% to 4.45\% with OpenAI weights.
N2L is higher on three pairs in $\bar A$, by at most 0.33\% (UCF B50Inc10 with LAION ViT-L/14, SUN B150Inc30 and CIFAR100 B50Inc10 with OpenAI ViT-B/32), and on ten pairs in $A_T$, by at most 0.54\%.
Seven of these ten are on Food and ImageNet-R with OpenAI weights, where the gain of {\ours} in $\bar A$ is also at most 0.61\%.

\subsection{Noisy Task Boundaries}
\label{supp:noisy}

Unlabeled images rarely respect task boundaries. We test whether {\ours} remains effective when the unlabeled data of a task also contain images of earlier classes.
Following the protocol of N2L~\cite{li2026n2l}, from the second task on, images drawn from the training sets of earlier tasks are added to the unlabeled images of task $t$ until they make up a fraction $p$ of them, with $p=20\%$ or $50\%$.
The learner is not told which images they are and can only name them among the classes of task $t$, so no correct label is available for them.
In a zero-base setting 9 of the 10 tasks are noisy, whereas in a half-base setting the clean first task holds half of the classes.
\Cref{tab:supp-noisy} compares {\ours} with RAIL~\cite{xu2024rail}, ENGINE~\cite{zhou2025engine} and N2L under this protocol at both noise levels.

{\ours} has the best $\bar A$ and $A_T$ on all 12 settings at both noise levels, and every Mean column ranks the four methods in the same order at both levels.
At $p=20\%$ its Mean gain over N2L in $\bar A$, 2.96\% and 3.15\% on the two blocks, is as large as without noise in \cref{tab:main}, where it is 3.19\% and 2.49\%.
At $p=50\%$ it shrinks to 1.77\% and 1.72\%, while the gain in $A_T$ remains 3.24\% and 2.23\%.
The drop comes mostly from the zero-base settings, where the gain in $\bar A$ falls from 2.83\% to 0.81\% while it stays at 2.68\% in the half-base settings.
Soft labels lie closer to the true labels than hard labels on the images of the current classes (\cref{supp:pl-quality}), and also on an image of an earlier class, for which no current class is correct, since a soft label $q_i$ is at squared distance $1+\|q_i\|^2$ from its true label against 2 for a hard label.
This lowers the target error that each such image brings into the classifier and is consistent with the unchanged lead at $p=20\%$.
From no noise to $p=50\%$, the Mean $\bar A$ of {\ours} falls by 3.28\% and 3.60\%, more than the 1.86\% and 2.83\% of N2L, while the falls in $A_T$, 4.19\% and 3.96\% against 3.43\% and 4.39\%, are of similar size.
The larger fall in $\bar A$ comes from the part of the gain that {\ours} draws from the images.
EM corrects the zero-shot distributions by fitting one component per current class to the images of a task (\cref{sec:em}) and is worth up to 4.18\% of $\bar A$ without noise (\cref{fig:ablation_posterior}), whereas the self-training of N2L adds little after its first round (\cref{supp:pl-acc}).
Images of earlier classes fit none of these components, and at $p=50\%$ in a zero-base setting they outnumber the images of each current class several times over, so they can pull the components away from the classes that EM should recover.
Such images therefore cost {\ours} more than they cost N2L, yet it stays above N2L on every setting.
These images fall outside the assumption that a task brings images of its own classes, and a component for them is left to future work.

\subsection{N2L with Components of PACIL}
\label{supp:transfer}

N2L~\cite{li2026n2l} uses the ridge coefficient $\lambda=0.1$ and a fixed fusion weight $\omega=0.8$, and its results in \cref{fig:n2l_variants} are our reproduction with its official code.
Two components of {\ours} carry over to N2L without its soft labels, $\lambda=10$ and the LOO choice of $\omega_t$ in \cref{eq:omega}, which then scores the LOO decisions with the hard pseudo-labels of N2L and keeps its sample weights.
We add them to this reproduction one at a time and together, to test whether the gain of {\ours} comes only from them.
Averaged over the 18 settings, $\lambda=10$ changes $\bar A$ and $A_T$ by 0.24\% and $-0.10\%$, and the LOO choice by 0.13\% and 0.36\%.
Neither helps everywhere, as $\lambda=10$ lowers $\bar A$ on both UCF settings, by up to 0.86\%, and the LOO choice gains most on ObjectNet, by 1.43\% and 1.11\% in $\bar A$.
Together they raise $\bar A$ and $A_T$ by 0.61\% and 0.64\%, more than the sum of their separate effects.
With $\lambda=10$ the leverages are smaller, so the LOO outputs amplify the fitted residuals less (\cref{supp:lambda}), and the selected $\omega_t$ rises from 0.85 to 0.91 on average.

{\ours} stays 1.63\% and 1.75\% above the strongest variant and is higher on 16 of the 18 settings in $\bar A$ and on 17 in $A_T$.
The two components thus explain about a quarter of the gain of {\ours} over N2L, and most of it comes from how {\ours} estimates and keeps the posterior of each label.
The remaining gap is largest on Aircraft, by 6.23\% and 6.73\% in $\bar A$ and by 7.72\% and 7.13\% in $A_T$, where N2L mislabels more than 40\% of the training images and the argmax of the soft labels of {\ours} alone brings N2L 37\% to 66\% of the gain in $\bar A$ (\cref{supp:pl-acc}).
On ObjectNet, the two components raise $\bar A$ of N2L the most, by 4.20\% and 1.93\%, and the strongest variant is level with {\ours} in $\bar A$ on B0Inc20 and above it on B100Inc20, by 0.36\% in $\bar A$ and 0.54\% in $A_T$.
ObjectNet B0Inc20 is also the only setting where {\ours} itself gains from a $\lambda$ above 10 (\cref{supp:lambda}).

\begin{figure}[!t]
\setlength{\abovecaptionskip}{0.2cm}
\setlength{\belowcaptionskip}{-0.2cm}
    \centering
	\begin{overpic}[width=\linewidth]{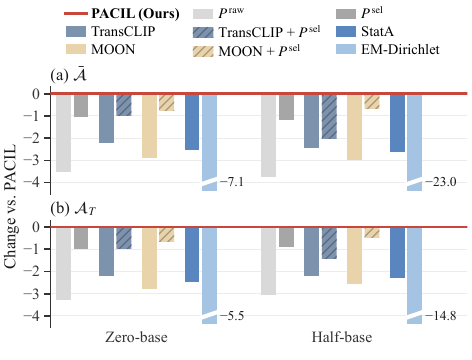}
    \end{overpic}
    \caption{Comparison with transductive methods.
Change in (a)~$\bar A$ and (b)~$A_T$ (\%) relative to {\ours} when its soft labels are replaced by $\Praw$ or $\Psel_t$ without EM or by the distributions of transductive methods, averaged over the nine datasets in the zero-base and the half-base settings.
All sources share the classifier and fusion of {\ours}, $+\Psel$ gives a method $\Psel_t$ as its zero-shot input, and the bars of EM-Dirichlet are labeled with their values.}
 \label{fig:label_sources}
\end{figure}

\subsection{Comparison with Transductive Methods}
\label{supp:transductive}

Transductive methods also estimate a class distribution for each unlabeled image, but only to predict these images (\cref{sec:related}).
We replace the soft labels $q_i$ of each task by the distributions that TransCLIP~\cite{zanella2024transclip}, StatA~\cite{zanella2025stata}, MOON~\cite{huang2026moon} and EM-Dirichlet~\cite{martin2024transductive} compute on $\mathcal{D}_t$ from the same features and class embeddings with their zero-shot hyperparameters, and keep the rest of {\ours} unchanged.
As references, we also train on $\Praw$ and $\Psel_t$ without EM, and on TransCLIP and MOON with $\Psel_t$ as their zero-shot input.

In \cref{fig:label_sources}, {\ours} is above all four methods in both metrics.
Averaged over the 18 settings, it is 2.34\% to 2.94\% above TransCLIP, StatA and MOON in $\bar A$ and 2.22\% to 2.69\% in $A_T$, and it is lower only on the two SUN settings and in $A_T$ on Food B50Inc10, by at most 0.31\%.
EM-Dirichlet falls below even $\Praw$, as it expects a batch to hold only a few classes.
The gap comes first from the class prior, as TransCLIP, StatA and MOON keep $\Praw$ as the prior of each image.
$\Psel_t$ alone, without EM, is above all four methods in both metrics, and as their zero-shot input it narrows the gap of TransCLIP and MOON to {\ours} from 2.34\% to 1.49\% and from 2.94\% to 0.73\% in $\bar A$.
The remaining gap lies mainly in the likelihood, as {\ours} fits vMF components to features whitened with the within-class scatter of all seen classes (\cref{sec:em}), whereas TransCLIP uses Gaussians with a shared diagonal covariance and MOON does not whiten.
Whitening with the statistics of the current task alone is enough to lower $\bar A$ of {\ours} by 0.37\%, which is half of its lead over MOON with $\Psel_t$.{\widowpenalty=10000\par}

\subsection{Pseudo-Label Quality}
\label{supp:pl-quality}

\paragraph{Accuracy.}
\label{supp:pl-acc}
\Cref{fig:pseudo_label_pacil_vs_n2l}a compares the pseudo-label accuracy of N2L~\cite{li2026n2l} at zero-shot and after each of its three refinement rounds with that of {\ours} on Aircraft and CUB, in the zero-base and half-base settings.
Accuracy is the fraction of training images whose label is the true class, pooled over all tasks of a setting, and ground-truth labels are used only for the measurements of this subsection.
{\ours} keeps a label distribution $q_i$ for each image, and to compare it with the hard labels of N2L we take its argmax.
This comparison is conservative for {\ours}, since its classifier is fitted to the whole $q_i$, and an image whose argmax is wrong keeps part of its probability mass on the true class. \looseness=-1

\begin{figure}[!t]
\setlength{\abovecaptionskip}{0.2cm}
    \centering
	\begin{overpic}[width=\linewidth]{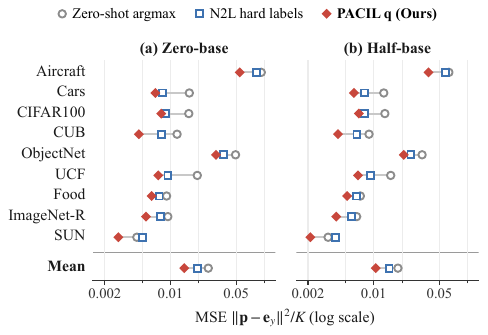}
    \end{overpic}
    \caption{Squared error of the pseudo-labels.
MSE of the zero-shot argmax (grey circles), the hard pseudo-labels of N2L~\cite{li2026n2l} after round 3 (blue squares) and the soft labels $q$ of {\ours} (red diamonds), averaged over the training images of all tasks of each setting in (a)~the zero-base and (b)~the half-base settings. The last row is the mean over the nine datasets. $K$ differs across datasets, so markers are compared within a row.}
 \label{fig:pseudo_label_mse}
\end{figure}

\begin{figure*}[!t]
\setlength{\abovecaptionskip}{0.2cm}
    \centering
	\begin{overpic}[width=\textwidth]{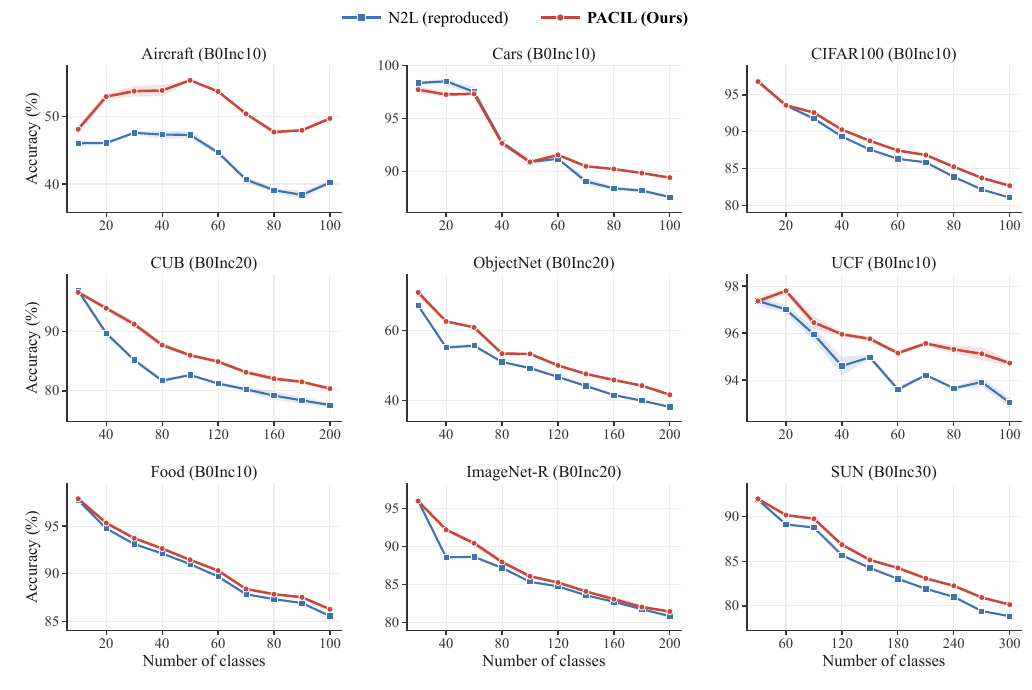}
    \end{overpic}
    \caption{Accuracy (\%) after each task in the zero-base settings, against the number of seen classes.
Accuracy is measured on the test images of all seen classes with the fused prediction.
N2L~\cite{li2026n2l} is re-run with its official code, since its paper reports no per-task accuracy.
Lines are means over three seeds, and shaded bands show standard deviation.}
 \label{fig:per_task_zero}
\end{figure*}

\begin{figure*}[!t]
\setlength{\abovecaptionskip}{0.2cm}
    \centering
	\begin{overpic}[width=\textwidth]{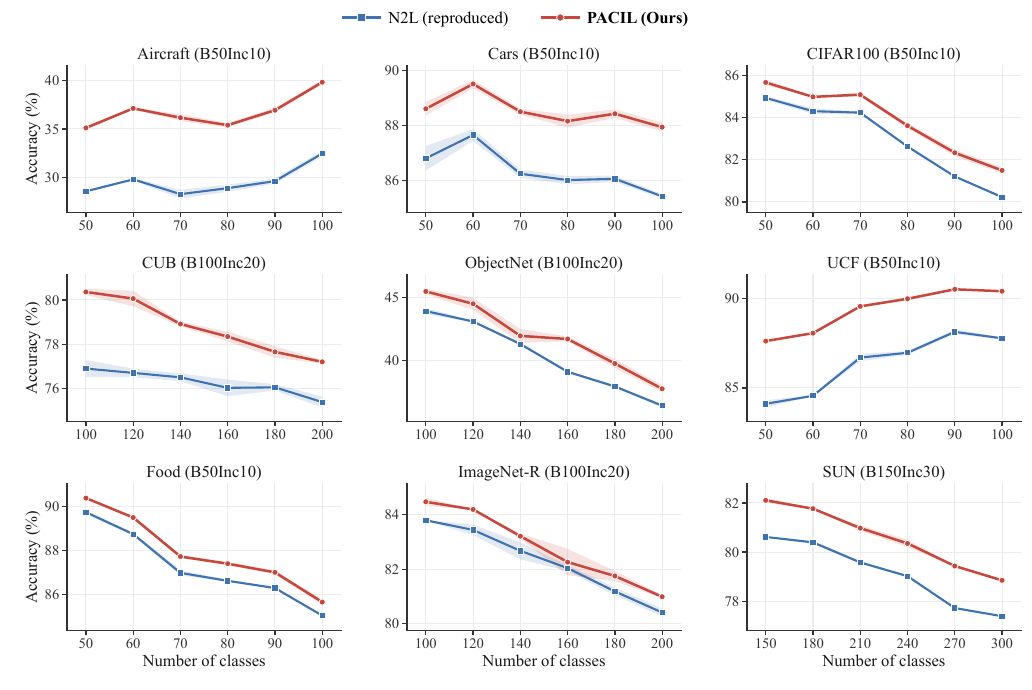}
    \end{overpic}
    \caption{Accuracy (\%) after each task in the half-base settings, where the first task holds half of the classes.
N2L~\cite{li2026n2l} is re-run with its official code, lines are means over three seeds, and shaded bands show standard deviation.}
 \label{fig:per_task_half}
\end{figure*}

Both methods start from the same zero-shot labels, and {\ours} is above N2L after round 3 on all four settings, by 9.73\% and 8.31\% on Aircraft and 2.78\% and 3.75\% on CUB.
N2L gains most in round 1, by up to 5.48\% on Aircraft and 3.14\% on CUB, and rounds 2 and 3 together add at most 0.59\%.
The margins are largest on Aircraft, where N2L is least accurate and mislabels more than 40\% of the images after round 3.
Self-training on hard labels corrects few errors after round 1 and can overwrite correct ones, while {\ours} corrects the class prior of the whole distribution and refines it by EM over the images of each task, which raises the accuracy on Aircraft from 53.85\% and 39.36\% at zero-shot to 69.00\% and 50.92\%. \looseness=-1

\Cref{fig:pseudo_label_pacil_vs_n2l}b separates the effect of these labels from the rest of {\ours}.
Replacing the pseudo-labels of N2L by the argmax of the {\ours} soft labels, with the rest of N2L unchanged, raises $\bar A$ by 3.58\% and 4.66\% on Aircraft and by 1.47\% and 1.34\% on CUB.
This is 37\% to 66\% of the gain of {\ours} in $\bar A$, and 47\% to 70\% in $A_T$.
The rest comes from the other differences between the two methods, such as training on the whole $q_i$ rather than its argmax and selecting $\omega_t$ on each task by closed-form LOO.

\paragraph{Squared error.}
\label{supp:pl-dist}
The bounds of \cref{thm:soft} depend on how close the targets of the classifier are to the label posteriors, which cannot be observed. \Cref{fig:pseudo_label_mse} therefore measures how far three targets are from the true labels.
They are the zero-shot argmax, which is also the starting label of N2L, the hard pseudo-labels of N2L after its three refinement rounds, and the soft labels $q_i$ of {\ours}.
A hard label enters as the one-hot vector of its class, and the distance of a target $p_i$ for an image $i$ of a task with $K$ classes is the mean squared error $\|p_i-e_{y_i}\|^2/K$, the Brier score averaged over these classes, so a hard label scores $2/K$ when it is wrong and 0 when it is right.
It is averaged over the training images of all tasks. \looseness=-1

The soft labels of {\ours} have the lowest MSE in all 18 settings.
Averaged over the 18 settings, their MSE is 0.0123, against 0.0169 for N2L and 0.0216 for the zero-shot argmax.
The refinement of N2L lowers the MSE of the zero-shot argmax by 22\% on average, but the MSE of $q_i$ is 9\% to 46\% below that of N2L in single settings.
SUN is the exception to this ordering, as the refinement there moves the hard labels away from the true labels, from 0.0044 to 0.0050 and from 0.0033 to 0.0039, in line with the accuracy of these labels, which falls from 93.42\% to 92.46\% and from 86.77\% to 85.70\% over the three rounds.

This ordering reflects the distance to the posteriors that \cref{thm:soft} needs.
For any target $p_i$ computed from the observations, the identity of \cref{sec:head} gives
\begin{equation}
\mathbb{E}_t\|p_i-e_{y_i}\|^2=\|p_i-\bar e_i\|^2+1-\|\bar e_i\|^2,
\label{eq:pl-brier}
\end{equation}
whose last two terms do not depend on the target.
The Brier score against the true labels therefore ranks targets by their squared distance to the posteriors, and its expectation is smallest at $p_i=\bar e_i$. In a zero-base setting every image has the same $K$, so the MSE ranks them in the same way. A hard label also stays at squared distance at least $(1-\bar e_{ic})^2$ from the posterior on every image by \cref{eq:onehot}, even when its class is correct, whereas a soft label can match the posterior. \Cref{fig:pseudo_label_mse} thus shows that $q_i$ comes closer to the posteriors than both kinds of hard labels, and this distance is what the bound of \cref{thm:soft}(a) depends on.

\begin{table}[!t]
\caption{Measured training and inference cost.
Means over the 18 settings with three runs each, on one A100 GPU with the same pre-extracted ViT-B/16 features.
Times are per setting and summed over its tasks.
Memory is the peak GPU memory during training, including the CLIP weights.
N2L inverts its $5000\times5000$ matrix on the CPU as in its official code, and N2L$^\dagger$ inverts it on the GPU.}
\label{tab:supp-cost}
\centering\footnotesize
\begin{tabular}{@{}lccc@{}}
\toprule
 & N2L & N2L$^\dagger$ & {\ours}\\
\midrule
Training time (s) & 22.61 & 2.81 & 2.16\\
\quad labeling & 2.36 & 1.90 & 1.18\\
\quad classifier update & 20.12 & 0.78 & 0.85\\
Inference time (s) & 7.61 & 7.33 & 7.33\\
Training memory (GB) & 1.45 & 2.54 & 3.23\\
\midrule
$\bar A$ (\%) & 75.64 & 75.64 & 77.92\\
$A_T$ (\%) & 72.32 & 72.32 & 74.81\\
\bottomrule
\end{tabular}
\end{table}

\subsection{Accuracy after Each Task}
\label{supp:per-task}

\Cref{fig:per_task_zero,fig:per_task_half} show the accuracy on all seen classes after each task, with N2L~\cite{li2026n2l} re-run from its official code.
In the zero-base settings, {\ours} is higher on 81 of the 90 tasks and level on 2, by 2.25\% on average.
It is lower only on the first four tasks of Cars, by at most 1.25\%, and on the first task of CIFAR100, CUB and ImageNet-R, by at most 0.27\%.
In the half-base settings it is higher on all 54 tasks, by 2.22\% on average and by at least 0.22\%.

Averaged over the datasets, the zero-base gap grows from 0.56\% after the first task to 2.61\% after the last, consistent with the errors of frozen pseudo-labels adding up over the tasks (\cref{thm:soft}).
In the half-base settings, the gap is 2.26\% after the first task, which holds half of the classes.
The gain is largest on the fine-grained Aircraft, 9.49\% and 7.32\% at the last task, where zero-shot CLIP~\cite{radford2021clip} has an $A_T$ of 17.16\% (\cref{tab:main}).
It is smallest on Food and ImageNet-R, 0.58\% to 0.71\%, where zero-shot CLIP reaches 81.99\% and 76.62\% (\cref{tab:supp-more}).
There the zero-shot argmax is mostly right, and a posterior concentrated on one class is close to its argmax by \cref{eq:onehot}, so keeping the whole distribution changes little.

\subsection{Training and Inference Cost}
\label{supp:cost}

\Cref{tab:supp-cost} reports the measured cost of training and inference.
All methods read the same pre-extracted CLIP features and run the 18 settings with three seeds each on one A100 GPU, so the comparison covers the learning step itself.
N2L~\cite{li2026n2l} is re-implemented here in our code base following its official implementation, which inverts the $5000\times5000$ matrix of its classifier with NumPy on the CPU, and N2L$^\dagger$ computes the same inverse on the GPU.

{\ours} takes 2.16\,s on average to train on one setting, of which 1.18\,s is labeling.
Labeling is cheap because each task runs EM only twice, once for the prior selection rule and once for the final run, on $d_p=128$ whitened dimensions, where one iteration costs $O(n_tk_td_p)$.
N2L takes 22.61\,s, almost all of it in the inverse on the CPU, since moving this step to the GPU reduces its classifier update from 20.12\,s to 0.78\,s and its training time to 2.81\,s.
The classifier update of {\ours} costs $O(n_tD^2)$ for the statistics and for the closed-form LOO, plus one $D\times D$ Cholesky factorization, the same order as the inverse of N2L$^\dagger$.
With the same kind of solver, the training time of {\ours} is therefore on par with that of N2L. \looseness=-1

At inference, both methods only add the random features and two matrix products, with the classifier and with the text embeddings, to the CLIP forward pass, and their inference time is the same, 7.33\,s per setting and about 0.32\,ms per image.
{\ours} keeps its statistics in double precision, so its peak training memory is 3.23\,GB on average against 2.54\,GB for N2L$^\dagger$. N2L itself uses only 1.45\,GB because its inverse runs on the CPU, outside the measured GPU memory. Across tasks, {\ours} carries only $(G,g,N,H,\nu)$ and $(A_t,C_t)$, which take $8(d^2+dK_t+K_t+d+1)+8(D^2+DK_t)$ bytes and do not grow with the number of seen images. They are dominated by the $D\times D$ matrix $A_t$, which N2L also keeps. \looseness=-1

\clearpage

\end{document}